%% file: main.tex
\documentclass[10pt,twocolumn,letterpaper]{article}

\usepackage[pagenumbers]{iccv} % To force page numbers, e.g. for an arXiv version


\usepackage{xcolor, pifont}
\usepackage{graphicx}
\usepackage{multirow}
\usepackage{bbding}
\usepackage{adjustbox}
\usepackage{appendix}
\usepackage{caption}
\newcommand{\tablestyle}[2]{\setlength{\tabcolsep}{#1}\renewcommand{\arraystretch}{#2}\centering\footnotesize}

\newcommand{\cmark}{\textcolor{green}{\ding{51}}} % 绿色的勾
\newcommand{\xmark}{\textcolor{red}{\ding{55}}}   % 红色的叉

\definecolor{iccvblue}{rgb}{0.21,0.49,0.74}
\definecolor{mygray}{gray}{.9}
\usepackage[pagebackref,breaklinks,colorlinks,allcolors=iccvblue]{hyperref}

\def\paperID{4875} % *** Enter the Paper ID here
\def\confName{ICCV}
\def\confYear{2025}

\title{HIS-GPT: Towards 3D Human-In-Scene Multimodal Understanding}

\author{
    {Jiahe Zhao$^{1,2}$, Ruibing Hou$^{1}$*, Zejie Tian$^{3}$, Hong Chang$^{1,2}$, Shiguang Shan$^{1,2}$} \\
    \small {$^1$Key Laboratory of AI Safety of Chinese Academy of Sciences (CAS), 
    Institute of Computing Technology, CAS, China} \\
    \small {$^2$University of CAS , China} \quad
    \small {$^3$Communication University of China}
    \\
    \small \texttt{zhaojiahe22@mails.ucas.ac.cn, \{houruibing, changhong, sgshan\}@ict.ac.cn} \\
    \small \texttt{2021211023003@mails.cuc.edu.cn}
}

\begin{document}

\twocolumn[{%
    \renewcommand\twocolumn[1][]{#1}%
    \maketitle
    \begin{center}
        \centering
        \captionsetup{type=figure}
        \includegraphics[width=0.9\linewidth]{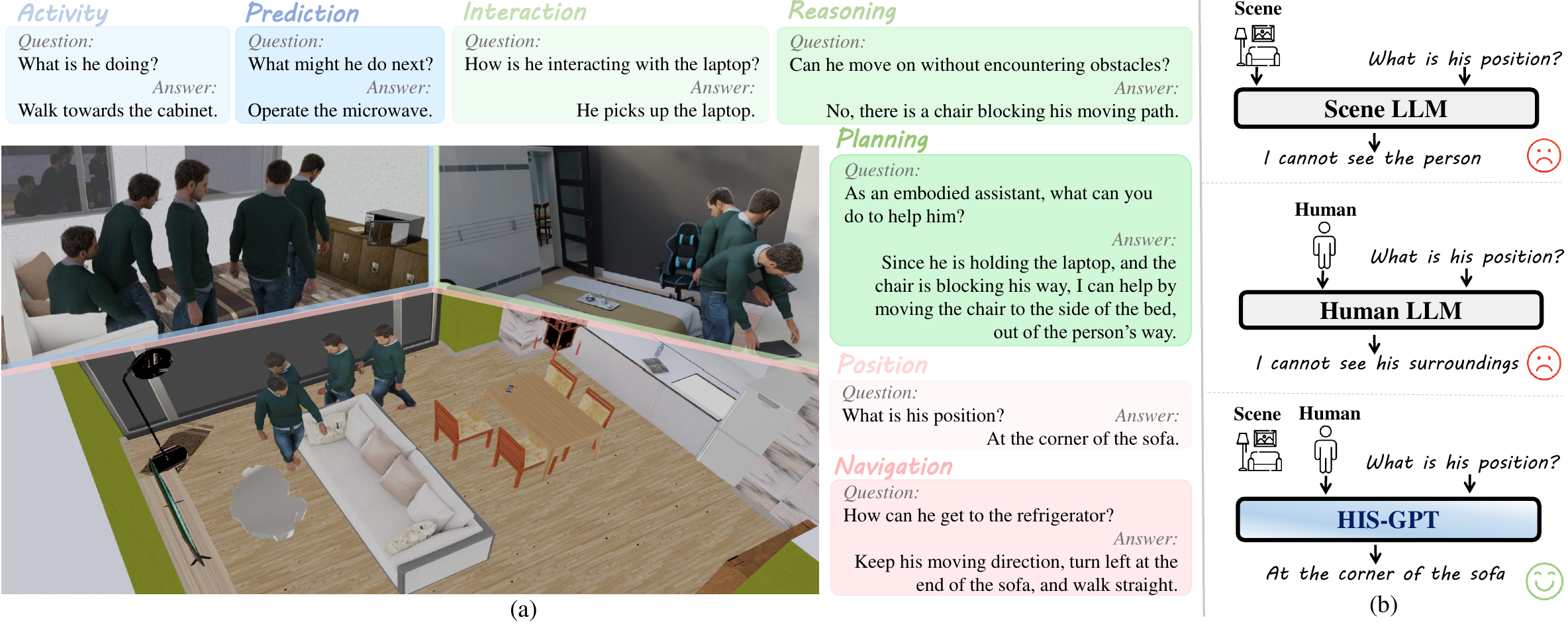} 
               \captionof{figure}{(a) \textbf{Illustration of HIS-QA task,} which  understands human behaviors in scene context. HIS-QA tasks span from basic perception tasks, such as recognizing human activity, interaction, and position in scene, to higher order functions like prediction, reasoning, planning, and navigation, facilitating embodied intelligence in real world. (b) \textbf{Illustration of HIS-GPT}. Unlike previous models that focus solely on either scene or human understanding, HIS-GPT could jointly perceive scene and human modalities to tackle the challenges of HIS-QA.}
        \label{fig:teaser}
    \end{center}%
}]

% \maketitle
\input{sec/0_abstract}
\let\thefootnote\relax\footnotetext{*Corresponding author.}
\input{sec/1_intro}
\input{sec/2_related-work}
\input{sec/3_his-bench}
\input{sec/4_his-gpt}

\input{sec/5_experiments}
\section{Conclusion}
\label{sec:conclusion}
In this paper, we introduce HIS-QA, a new task formulation for Human-In-Scene understanding. To evaluate this task, we raise HIS-Bench, the first multimodal benchmark tailored for HIS-QA, featuring diverse questions that span basic perception, complex reasoning and embodied applications. Additionally, we propose HIS-GPT, a foundation model that jointly perceives 3D human and scene inputs, effectively addressing HIS-QA tasks in a unified framework. We believe the benchmark and model could benefit future research and applications in human-centric understanding.

\noindent{\textbf{Acknowledgements.}} This work is partially supported by National Natural Science Foundation of China U2336213, 62376259, 62306301.

\input{sec/X_suppl}

{
    \small
    \bibliographystyle{ieeenat_fullname}
    \bibliography{main}
}

\end{document}

%% file: sec/0_abstract.tex
\begin{abstract}
We propose a new task to benchmark human-in-scene understanding for embodied agents: Human-In-Scene Question Answering (HIS-QA).  Given a human motion within a 3D scene, HIS-QA requires the agent to comprehend human states and behaviors, reason about its surrounding environment, and answer human-related questions within the scene. To support this new task, we present HIS-Bench, a multimodal benchmark that systematically evaluates HIS understanding across a broad spectrum,  from basic perception to commonsense reasoning and planning.   Our evaluation of various vision-language models on HIS-Bench reveals significant limitations in their ability to handle HIS-QA tasks. 
To this end, we propose HIS-GPT, the first foundation model for HIS understanding. HIS-GPT integrates 3D scene context and human motion dynamics into large language models while incorporating specialized mechanisms to capture human-scene interactions. Extensive experiments demonstrate that HIS-GPT sets a new state-of-the-art on HIS-QA tasks. We hope this work inspires future research on human behavior analysis in 3D scenes, advancing embodied AI and world models. Codes and data will be available at \href{https://github.com/ZJHTerry18/HumanInScene}{https://github.com/ZJHTerry18/HumanInScene}.
\end{abstract}

%% file: sec/1_intro.tex
\section{Introduction}
\label{sec:intro}

%% Development in 3D scene & 3D human MLLMs
In recent years, intelligent systems for 3D vision-language understanding have witnessed remarkable progress~\cite{qi2024gpt4point,chen2024grounded,gu2024conceptgraphs,yin2025shapegpt,qi2024shapellm,sun20233d}, largely driven by the advancements in Large Language Models (LLMs)~\cite{chiang2023vicuna,dubey2024llama,chowdhery2023palm,zhang2022opt,chung2024scaling}. Specifically, 3D scene LLMs~\cite{fu2024scene,huang2024chat,hong20233d,yang2024regionplc} excel in tasks such as captioning and grounding within 3D layouts, whereas 3D human LLMs~\cite{jiang2023motiongpt, zhou2024avatargpt, li2024unipose, luo2024m3gpt} exhibit strong capabilities in open-ended interpretations of human poses and motions. By embracing 3D world, these models significantly promote the developments in robotics and embodied AI. 

%% Limitations in current 3D understanding, the reason to propose Human-in-scene task
Despite significant progress in separately perceiving 3D scenes and humans, a critical yet underexplored challenge remains: \textbf{human-in-scene (HIS)} understanding. This task requires an agent to jointly comprehend human subjects and their surrounding environments to capture intricate interactions and relationships. 
Such capability is essential for accurately recognizing fundamental human states (\eg, \textit{positioned in front of the TV)} and actions (\eg, \textit{sit on a chair}) in real-world scenarios. With effective HIS understanding, embodied agents could reason, predict, and react based on their observations of human-scene dynamics, thereby serving as versatile assistants in applications such as household robots. However, the current limitations of 3D LLMs to integrate human and scene perception largely hinder further advancements in embodied intelligence.

%% The introduction of HIS-QA task. The proposal of HIS-Bench. Task converages.
To bridge this critical gap, we introduce~\textbf{HIS-QA}, a novel task for Human-In-Scene Question Answering,  where an agent answers questions about human states and behaviors within a 3D scene, as depicted in~\cref{fig:teaser} (a).  %HIS-QA is formulated as follows: given a human motion within a 3D scene, the agent needs to answer questions regarding human states or behaviors in relation to the scene context. 
To systematically evaluate this task, we propose \textbf{HIS-Bench}, the first multimodal benchmark tailored for HIS understanding. As shown in~\cref{tab:benchmark compare}, HIS-Bench differs from previous benchmarks by integrating both human and scene modalities for open-ended, language-guided understanding.
A major challenge in constructing HIS-Bench is the lack of detailed textual annotations in existing HIS datasets~\cite{hassan2019resolving, hassan2021stochastic, araujo2023circle, zheng2022gimo, cong2024laserhuman, jiang2024scaling}, which primarily provide coarse action labels (\eg, \textit{walking}, \textit{sitting}). Additionally, the intrinsic 3D spatial complexity of human-scene interactions makes it impractical to generate precise annotations using proprietary models like GPT-4o~\cite{hurst2024gpt}.
To overcome this limitation, we develop a specialized data annotation pipeline %that produces diverse and detailed linguistic descriptions for HIS data. As shown in Figure XX, this pipeline 
that combines advanced 3D understanding tools with rule-based algorithms for text annotations. This pipeline enables the generation of rich annotations covering human actions, scene properties, and human-scene interactions. 
Built upon these detailed annotations, HIS-Bench comprises 800 questions organized hierarchically into 3 general abilities, 7 core tasks, and 16 sub-tasks, spanning a broad spectrum from basic human activity perception to advanced reasoning, prediction, and planning. This comprehensive benchmark establishes a new standard for evaluating HIS understanding. 

%% Evaluation on self-crafted baselines.
Utilizing  HIS-Bench, we systematically evaluate HIS-QA with existing vision-language models~\cite{li2024llava, bai2025qwen2, hurst2024gpt, huang2024chat}. We observe that existing models fall short in HIS understanding, largely due to their insufficient capacity for jointly modeling human-scene characteristics in 3D space. 
To address the above limitation, we propose \textbf{HIS-GPT}, a multimodal large language model tailored for HIS understanding.  As shown in~\cref{fig:teaser} (b), HIS-GPT fundamentally  differs from prior 3D LLMs~\cite{hong20233d, chen2024ll3da, huang2024chat} by jointly interpreting 3D scenes and humans.  Specifically, HIS-GPT integrates a scene encoder~\cite{zhou2023uni3d} and a motion encoder~\cite{luo2024m3gpt} to extract structured representations of 3D environments and human motions. These representations are subsequently processed by the core LLM~\cite{chiang2023vicuna}, enabling seamless fusion of scene and motion cues to enhance capabilities on HIS tasks.  

Beyond previous 3D LLMs that focus on perceiving a single modality (either human or scene), a key challenge in HIS understanding lies in accurately modeling human-scene interactions. To this end, HIS-GPT introduces two critical components. On one hand,  an \textbf{Auxiliary Interaction (AInt)} module enhances interactive cues within each modality, through incorporating multiple training objectives that require a joint understanding of human and their surroundings. %These objectives include activity classification, spatial relation detection and contact detection.
By enforcing these constraints,  HIS-GPT is guided to learn enriched, contextually aware representations of human-scene interactions.
On the other hand, a \textbf{Layout-Trajectory Position Encoding (LTP)} module generates position embeddings by encoding the spatial distribution of major objects in the scene layout, along with the temporal trajectories of human motion at each timestamp. By infusing fine-grained spatiotemporal knowledge into latent representations of scene and human, LTP module enhances both modalities, effectively capturing the dynamic interplay between human motions and 3D environments. 

To our knowledge, HIS-GPT is the first approach to address the tasks of human-in-scene understanding.  
% To train HIS-GPT, we employ our annotation pipeline to generate 700K mulltimodal data covering over 700 3D scenes. 
Extensive experiments demonstrate that HIS-GPT achieves state-of-the-art performance on HIS-QA task, establishing a strong foundation for future research.

%% Summary of contributions:
%%% 1. introduce HIS-QA task
%%% 2. curate the first benchmark, HIS-Bench. Text annotation pipeline.
%%% 3. Evaluations on several off-the-shelf baselines
%%% 4. Build new foundation model, HIS-GPT.
% We summarize our contributions as follows:
% \begin{itemize}
%     \item We introduce a new task formulation of HIS-QA, designed to advance the study of intelligent agents in the challenging domain of HIS understanding, shedding inspirations on the development of embodied AI.
%     \item We propose HIS-Bench, a comprehensive benchmark featuring 800 high-quality questions, spanning 16 tasks from basic perception to complex reasoning and planning. An automatic annotation pipeline is devised to generate multi-perspective data for the benchmark.
%     \item Using existing expert models, we craft a set of baselines for HIS-QA, and conduct a wide range of evaluations on HIS-Bench, revealing the deficiencies of current approaches on tackling HIS-QA task.
%     \item We propose HIS-GPT, the first foundation model for human-in-scene understanding. Benefiting from innovative human-scene interaction modules and over 550k self-curated data, HIS-GPT achieves state-of-the-art performance on HIS-QA task, serving as a strong base framework of future research.
% \end{itemize}

%% file: sec/2_related-work.tex
\section{Related Work}
\label{sec:related work}

\begin{figure*}
    \centering
    \includegraphics[width=0.95\linewidth]{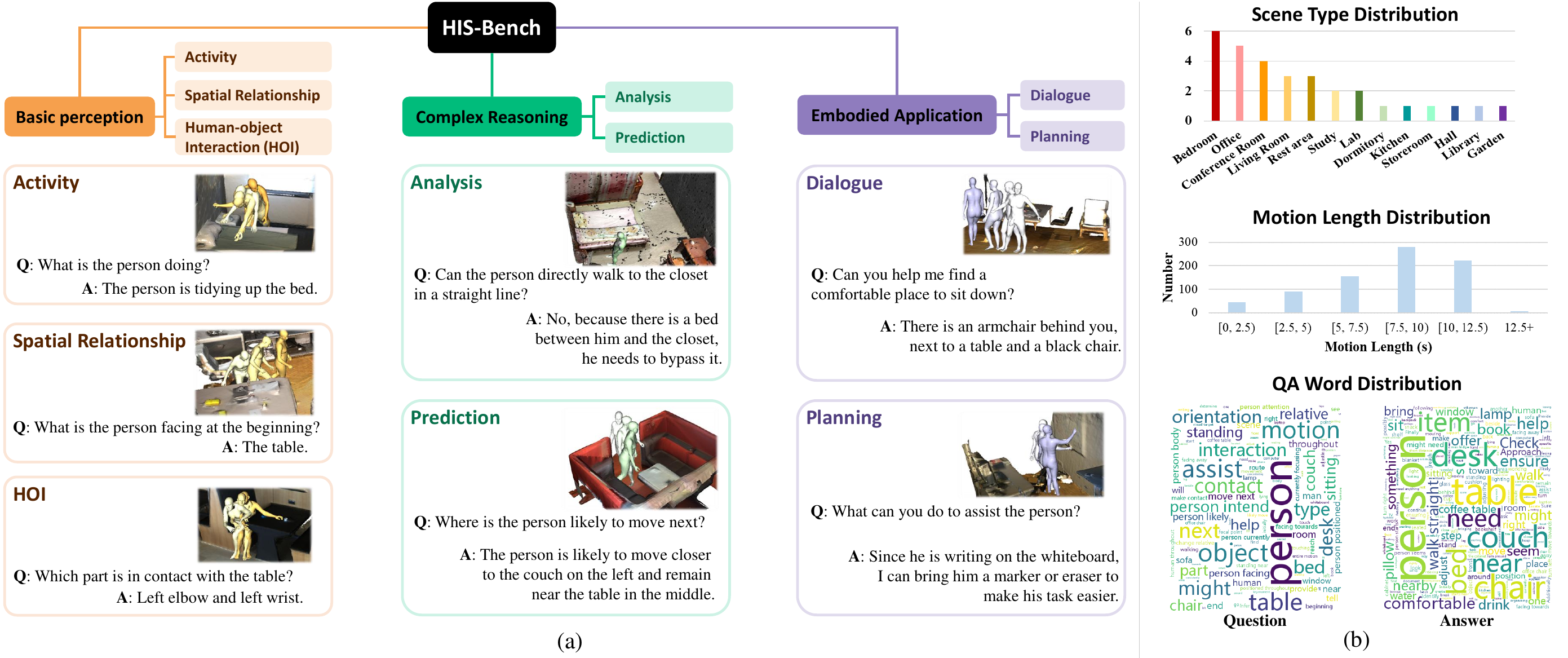}
     \vspace{-3mm}
     \caption{(a) \textbf{Task taxonomy and data samples of HIS-Bench.} HIS-Bench is structured into 3 general abilities and 7 core tasks. (b) \textbf{Statistics of HIS-Bench.} HIS-Bench is diverse in term of scene types, motion lengths, and word distributions.}
    \label{fig:his-bench}
    \vspace{-5mm}
\end{figure*}

\noindent\textbf{3D Scene-Language Understanding.} \ 3D scene-language understanding is a critical technique for agents to interact with the real world. It contains a wide range of tasks, including 3D captioning~\cite{chen2021scan2cap, chen2023end}, 3D visual grounding~\cite{zhang2023multi3drefer, achlioptas2020referit3d, yang20243d}, and 3D question answering~\cite{azuma2022scanqa, parelli2023clip}. Recent approaches adopt LLMs to tackle various 3D scene understanding tasks within a unified framework~\cite{hong20233d, hong2024multiply, yang2024regionplc, fu2024scene, huang2024chat, chen2024ll3da, zhen20243d}, benefiting from the synergies of multi-task learning. 

Despite their success in interpreting 3D scenes, these models are confined to tasks centered solely on scenes, and cannot handle 3D environments that include human elements. Some efforts~\cite{ma2022sqa3d, wang2024embodiedscan, zhang2024spartun3d} explore situated scene understanding by assuming the presence of a subject in 3D scenes. However, these approaches rely on explicit text inputs or first-person views to establish a subject's location, while also lacking full-body pose representation. In contrast, our proposed HIS-QA requires to directly model both 3D scene and humans from vision modalities, while being aware of the human pose. This setting allows for a more comprehensive perception of human states within the scene.

\input{tables/benchmark-compare}

\noindent\textbf{3D Human-Language Understanding.} \ 3D human-language understanding primarily focuses on recognizing human poses or motions~\cite{delmas2022posescript, delmas2023posefix, li2024unimotion, guo2022tm2t}. %since these represenations encompass rich geometric information. 
Recently, several works introduce LLMs to interpret human pose and motions~\cite{jiang2023motiongpt, zhou2024avatargpt, chen2024motionllm, zhang2024large, li2024unipose, luo2024m3gpt, feng2024chatpose}, addressing tasks like motion captioning~\cite{lin2023motion, guo2022generating} and question-answering~\cite{chen2024motionllm, endo2023motion}. However, these approaches overlook the environmental context of humans, constraining their ability to comprehensively recognize human status. To overcome this limitation, we present HIS-GPT, which processes human motions alongside scene contexts, enabling a more comprehensive understanding of human behavior in real-world environments.

%% file: tables/benchmark-compare.tex
\begin{table}
\centering
% \caption{Overview of existing benchmarks related to 3D scene and human. HIS-Bench is the first benchmark on natural language-guided, open-ended understanding of both human and scenes, with various text generation methods. `mo.gen.', `det.', `cap.' and `q.a.' refers to motion generation, detection, caption, and question-answering, respectively.}

\caption{Overview of existing benchmarks related to 3D scene and human. %HIS-Bench is the first benchmark designed for natural language-guided, open-ended understanding of both human and scenes.
`mo.gen.', `det.', `cap.' and `q.a.' refers to motion generation, detection, caption, and question-answering, respectively.}
\vspace{-2mm}
\begin{adjustbox}{width=\linewidth,center}
\tablestyle{5pt}{1.0}
\begin{tabular}{l|c|ccc|cc}
\toprule
    \multirow{2}{*}{Benchmark} & \multirow{2}{*}{Task} & \multicolumn{3}{c|}{Modalities} & \multicolumn{2}{c}{Language task} \\
    && Scene & Human & Language & Open-ended & Text generation \\
\midrule
    TRUMANS~\cite{jiang2024scaling} & mo.gen. & \cmark & \cmark & \xmark & - & - \\
\midrule
    ScanRefer~\cite{chen2020scanrefer} & det. & \cmark & \xmark & \cmark & \xmark & template \\
    SQA3D~\cite{ma2022sqa3d} & q.a. & \cmark & \xmark & \cmark & \xmark & template \\
    OpenEQA~\cite{majumdar2024openeqa} & q.a. & \cmark & \xmark & \cmark & \cmark & human \\
\midrule
    Motion-X~\cite{lin2023motion} & cap. & \xmark & \cmark & \cmark & \cmark & auto \\
    MoVid-Bench~\cite{chen2024motionllm} & q.a. & \xmark & \cmark & \cmark & \cmark & auto \\
\midrule
    \textbf{HIS-Bench(Ours)} & q.a. & \cmark & \cmark & \cmark & \cmark & human\&auto \\
\bottomrule
    
\end{tabular}
\end{adjustbox}
\label{tab:benchmark compare}
\vspace{-5mm}
\end{table}

%% file: sec/3_his-bench.tex
\section{HIS-Bench}
\label{sec:his-bench}

To explore the problem of understanding human behaviors in 3D scenarios,  we propose HIS-QA, a new task for addressing human-in-scene understanding of AI agents.  
% Given a human motion within a 3D scene, HIS-QA requires the agent to comprehend human states and behaviors, reason about the surrounding environment, and answer questions related to human in the scene. 
A problem instance in HIS-QA can be formulated as a quadruplet $\left\langle \mathcal{S}, \mathcal{M}, \mathcal{Q}, \mathcal{A} \right\rangle$. $\mathcal{S}$ denotes 3D scene in point cloud.
$\mathcal{M}$ denotes 3D human motion sequence, with each frame characterized by a SMPL pose~\cite{loper2015smpl}. 
$\mathcal{Q}$ refers to a natural language question and $\mathcal{A}$ is the ground-truth answer. The agent is tasked with generating an answer $\hat{\mathcal{A}} = \mathrm{Agent}(\mathcal{S}, \mathcal{M}, \mathcal{Q})$ that closely aligns with the true answer $\mathcal{A}$.

 However, existing 3D scene QA~\cite{azuma2022scanqa, ma2022sqa3d} and 3D human QA benchmarks~\cite{endo2023motion, chen2024motionllm} focus solely on scene or human understanding in isolation, overlooking human-scene interactions. To address this gap, we propose HIS-Bench, the first dedicated benchmark for HIS-QA. Next, we introduce the task taxonomy and data generation pipeline for HIS-Bench.  More details on constructing HIS-Bench are provided in Appendix~\ref{sec:supp-hisbench details}, and additional examples of HIS-Bench are provided in Appendix~\ref{sec:supp-hisbench samples}.

\subsection{Task Taxonomy}
\label{sec:his-bench_task taxonomy}
As shown in Fig. \ref{fig:his-bench}, HIS-Bench defines a structured taxonomy of benchmark tasks, encompassing three fundamental abilities: \textbf{\textit{basic perception}}, \textbf{\textit{complex reasoning}} and \textbf{\textit{embodied applications}}. 
These categories comprise 7 core tasks, further divided into 16 sub-tasks:

\begin{itemize}
    \item \textbf{Activity.} (1) \textit{Single Activity:} Recognize the human activity  within the scene. (2) \textit{Sequential Activity:} Recognize the human activity before or after a specific action.
    \item \textbf{Spatial Relationship.} (3) \textit{Human Position:} Identify the human's precise location in the scene. (4) \textit{Body Orientation:} Identify the human body's orientation relative to the scene. (5) \textit{Object Orientation:} Identify the object that is at a given orientation relative to the human.
    \item \textbf{Human-object Interaction.} (6) \textit{Interaction Type:} Recognize the type of human-object interaction. (7) \textit{Interacting Object:} Recognize the object the human is interacting with. (8) \textit{Contact Part:} List the human body parts in contact with a given object.
    \item \textbf{Analysis.} (9) \textit{Focus Analysis:} Infer the object or area the human is attending to. (10) \textit{Situated Analysis:} Deduce scene-related knowledge from the human's perspective, such as affordance and approachability.
    \item \textbf{Prediction.} (11) \textit{Intent Prediction:} Predict the human's next intended activity. (12) \textit{Movement Prediction:} Predict the human's future trajectories and spatial positions.
    \item \textbf{Dialogue.} (13) \textit{Situated Dialogue:} Complete a conversation with the human regarding the scene context.
    \item \textbf{Planning.} (14) \textit{High-level Task:} Provide a general plan to assist the human based on their status. (15) \textit{Low-level Task:} Provide step-by-step instructions for assisting the human. (16) \textit{Navigation:} Provide a route to guide the human towards a specified destination.
\end{itemize}

\begin{figure}
    \centering
    \includegraphics[width=0.95\linewidth]{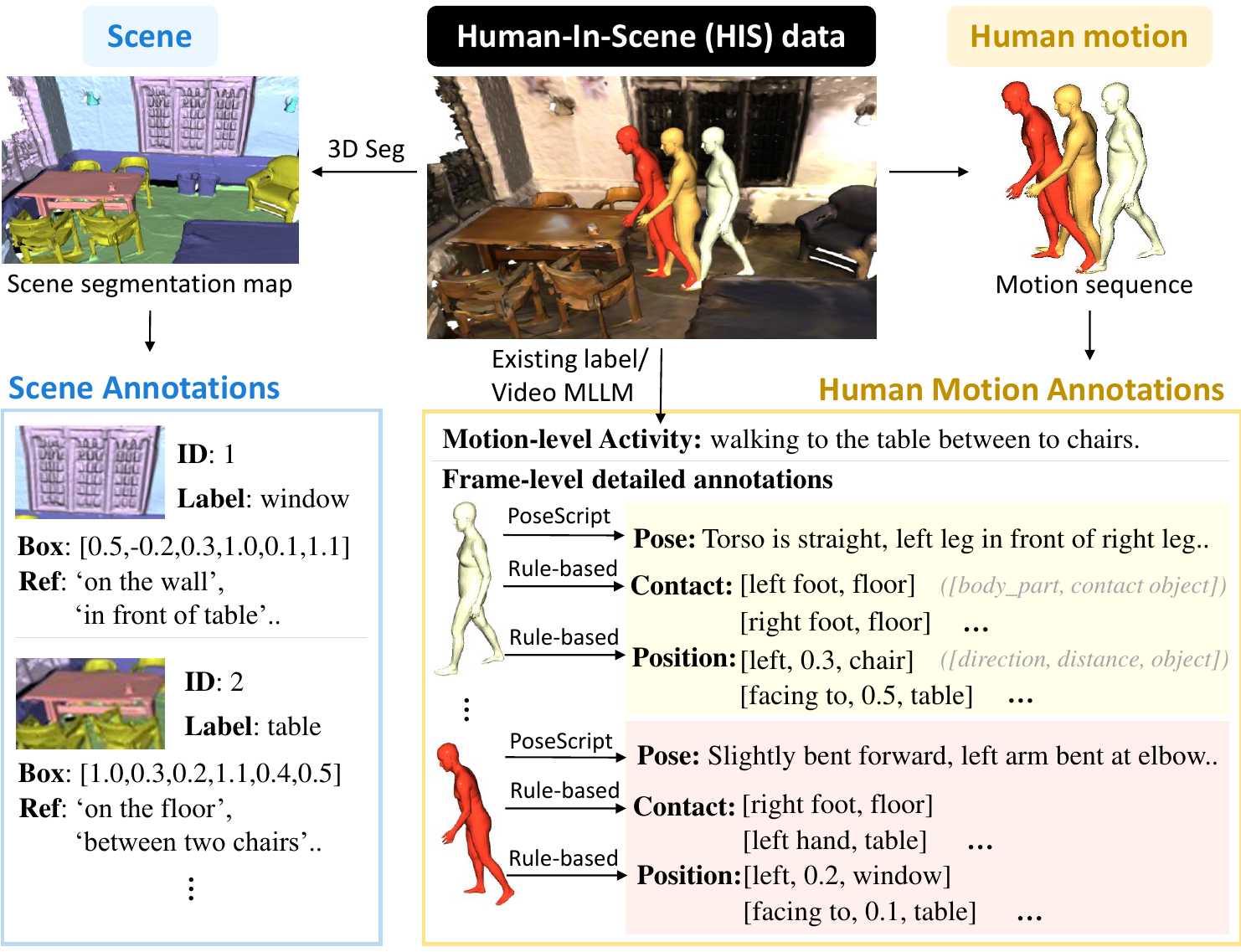}
    % \caption{Text annotation pipeline for HIS data. For scene annotations, we segment the 3D scene and derive instance-level labels, bounding boxes and reference expressions. For motion annotations, we get motion-level activities from existing labels or video MLLMs, while using expert models and rules to derive frame-level annotations on pose, human-scene contact and human position.}
      \caption{Text annotation pipeline for HIS data. For scene annotations, we segment the 3D scene to derive instance-level labels, bounding boxes, and reference expressions. For motion annotations, we obtain motion-level activities from existing labels or video MLLMs. Additionally, expert models and rules are used to generate frame-level annotations, including pose, human-scene contact, and human position.}
    \label{fig:data generation}
     \vspace{-5mm}
\end{figure}

\subsection{Data Generation Pipeline}
\label{sec:his-bench_data generation} 

\noindent\textbf{Text Annotation.} \ 
Acquiring multimodal resources for 3D HIS data is challenging, as existing datasets primarily contain 3D scene-language ~\cite{fu2024scene, yang20243d} or human-language~\cite{lin2023motion, mahmood2019amass}, but lack the necessary human-in-scene descriptions essential for HIS understanding.
To bridge this gap, we propose a multi-faceted annotation pipeline that generates rich and comprehensive human-scene descriptions. 

As shown in~\cref{fig:data generation}, our multi-faceted annotation pipeline comprises scene annotations and human motion annotations. For \textbf{Scene  Annotations}, following~\cite{fu2024scene}, we utilize 3D scene segmentation tools~\cite{schult2023mask3d} and visual caption models~\cite{li2023blip} to generate semantic labels, 6D bounding boxes, and referring expressions for key objects in the scene. 
For \textbf{Human Motion Annotations}, we first generate \textit{motion-level activities}: for scene data with recorded videos, a video captioner~\cite{li2024llava} is prompted to generate descriptions on human activities. For datasets lacking video recordings, we directly adopt the action labels provided in the original annotations. 
Additionally, we generate \textit{frame-level detailed annotations} for key frames in the motion sequence, including:  (1) \textit{Pose}: PoseScript~\cite{delmas2022posescript} is used to generate detailed narrations on part-level body postures. (2) \textit{Contact}: Utilizing SMPL fitting model~\cite{loper2015smpl}, we extract human joint locations and annotate those that establish contact with the 3D mesh of scene objects. (3) \textit{Position}: We design a rule-based approach to compute object orientation and distance relative to the human, categorizing these spatial relationships into predefined classes in natural language format. 

\begin{figure*}[!t]
    \centering
    \includegraphics[width=0.9\linewidth]{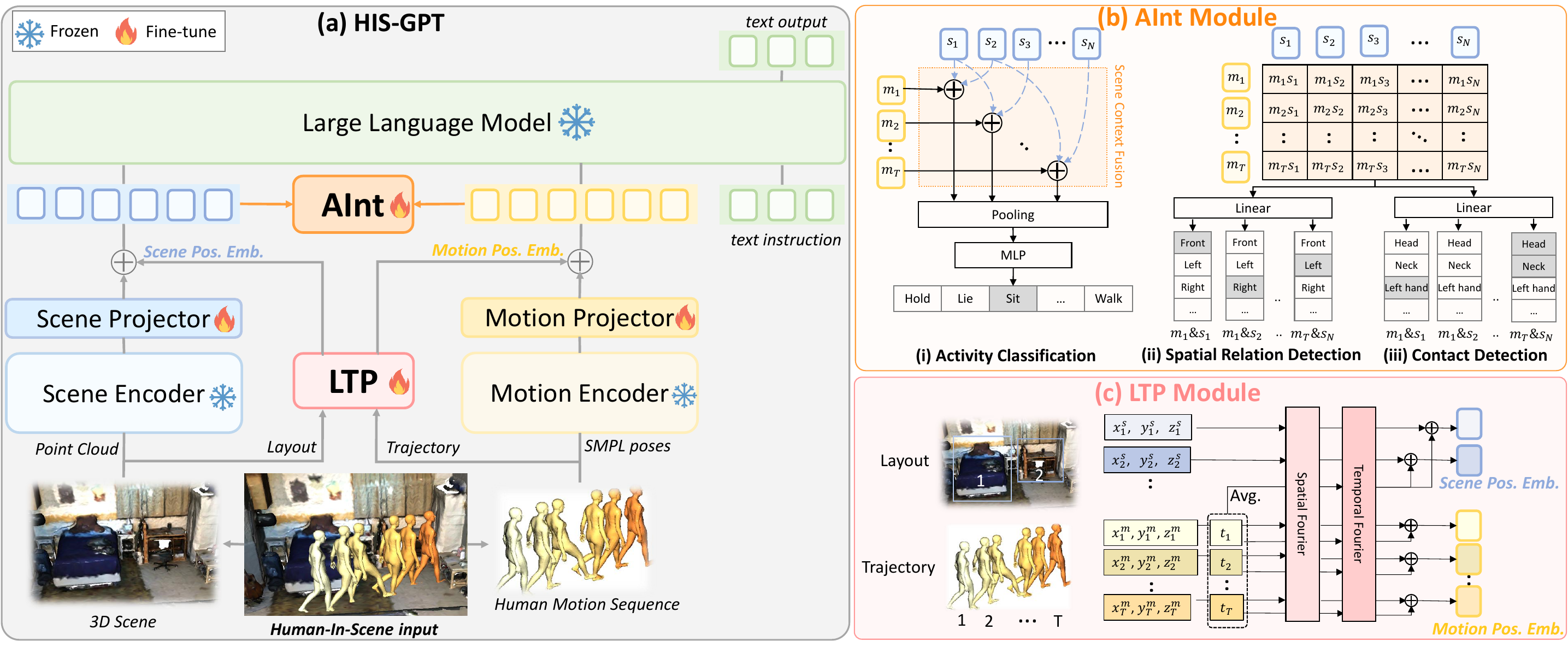}
    \caption{(a) \textbf{HIS-GPT overall architecture}. HIS-GPT uses separate pretrained encoders for scene and motion to extract embeddings, which are then combined with instructions and processed by the LLM. (b) \textbf{Auxiliary Interaction (AInt) module}: Enhance human-scene interactions through three auxillary  sub-tasks. (c) \textbf{Layout-Trajectory Position Encoding (LTP) module}: Encode spatial and temporal relationships into position embeddings, injecting contextual knowledge to enhance HIS understanding.}
    \label{fig:his-gpt}
    \vspace{-5mm}
\end{figure*}

\noindent\textbf{Benchmark Construction.} \ First, we collect 3D HIS data from PROX~\cite{hassan2019resolving} and GIMO~\cite{zheng2022gimo}, two high-quality HIS datasets covering diverse scenarios and human activities. Then, we apply our multi-faceted text annotation pipeline to generate linguistic labels, which are then fed into GPT~\cite{achiam2023gpt} with self-crafted prompts to create multiplex question-answer (QA) pairs, forming the foundation of HIS-Bench. 
% We give 1-2 in context examples for each task to help GPT learn the desired QA content. Specially, to ensure the answer quality for prediction tasks, we provide the subsequent motion in the original data as the reference for generating answers. 
 This process enables the construction of samples for 13 out of 16 sub-tasks. However, for focus analysis, situated analysis, and navigation tasks, existing annotations are insufficient. So we recruit human annotators to manually label these data. To ensure data quality, we manually verify each sample to preclude incorrectness or ambiguity in answers. After these procedures, we finalize HIS-Bench with 800 unique questions (each sub-task has 50 questions) covering 31 scenes and 500 motion segments, possessing diversity across scene types, motion patterns, and linguistic expressions. The statistics of HIS-Bench is presented in~\cref{fig:his-bench} (b).

%% file: sec/4_his-gpt.tex
\section{HIS-GPT}
\label{sec:his-gpt}
Existing vision-language models~\cite{huang2024chat, chen2024ll3da, chen2024motionllm} struggle to jointly model 3D human and scene modalities, limiting their effectiveness in HIS understanding. In this work, we propose HIS-GPT, a multi-modal framework designed to integrate human motion with scene context information, enabling more comprehensive HIS understanding.
%This section presents the overview of  model architecture and training procedure of HIS-GPT.

\subsection{Model Architecture}

\textbf{Overview.}  \ 
As shown in~\cref{fig:his-gpt}, HIS-GPT takes as input a 3D scene $\mathcal{S}$, a human motion sequence $\mathcal{M}$ and a text instruction $\mathcal{I}$. The scene is represented as a point cloud  $\mathcal{S} \in \mathbb{R}^{P \times 6}$, with each point characterized by 3D coordinates and RGB values. The  motion $\mathcal{M}=\left\{ M_i \right\}_{i=1}^{T}$ is a sequence of $T$ SMPL human poses. 
The 3D scene $\mathcal{S}$ and human motion $\mathcal{M}$ are encoded separately into latent embeddings using dedicated encoders. To enhance human-scene interactions, we introduce two key modules: the \textbf{Auxiliary Interaction (AInt)} module, which injects interaction-aware knowledge into the scene and motion embeddings, and the \textbf{Layout-Trajectory Position Encoding (LTP)} module, which encodes spatial and temporal relationships between scene and human motions. Finally, the enriched embeddings from both modalities are projected and prefixed to the text instruction $\mathcal{I}$, before being fed into the LLM to generate natural language answers. 

\noindent\textbf{Scene Encoder.} \ Following~\cite{huang2024chat}, we extract object features using a pretrained 3D encoder~\cite{zhou2023uni3d}, with object point clouds derived from a 3D scene segmentor~\cite{schult2023mask3d}.  The scene encoder generates a set of scene embeddings $\left\{s_i \in \mathbb{R}^d\right\}_{i=1}^{N}$ for 3D scene $\mathcal{S}$, where $N$ denotes the number of detected objects and $d$ is the latent embedding dimension.

\noindent\textbf{Motion Encoder.} \ Following~\cite{luo2024m3gpt}, we adopt a motion VQ-VAE~\cite{van2017neural} as the motion encoder. The motion encoder maps human motion $\mathcal{M}$ to a set of motion embeddings $\left\{ m_t \in \mathbb{R}^d \right\}_{t=1}^{T}$ derived from the learned motion codebook.

\noindent\textbf{Auxiliary Interaction (AInt) Module.} \ However, the scene and human motion embeddings are generated individually, lacking essential human-scene interactive cues. To address this, we propose AInt, which incorporates a set of auxiliary tasks to guide scene and motion embeddings in capturing these interactive cues, as shown in~\cref{fig:his-gpt} (b):

(1) \textit{Activity Classification.} As human activities involve interactions with surrounding scenes, we introduce an \textit{activity classification} task to predict human activity within the scene. In detail, we first perform scene context fusion by integrating motion embeddings with the features of objects likely to be involved in the activity. Specifically, for the motion $m_t$, we identify the $k$ nearest objects based on spatial proximity  to $m_t$, and fuse their latent embeddings with the motion embedding:
    
\vspace{-2mm}
\begin{small}
\begin{equation}
    \tilde{m}_{t} = m_{t} + \mathrm{Avg}\left(s_{t_1}, ..., s_{t_k}\right),
\end{equation}
\end{small}where $t_1 \sim t_k$ denotes the indices of the $k$ nearest objects for $m_t$, and $\mathrm{Avg}(\cdot)$ is the averaging operation.  The fused motion embedding is then passed through a  multi-layer perceptron (MLP) to predict the human activity category, supervised by a cross-entropy loss:

%   \begin{small}
% \begin{equation}
%     \mathcal{L}_{act} = \mathrm{CE}\left(p^a, \hat{p}^a\right), \
%     \hat{p}^a = \mathrm{MLP}(\mathrm{Avg}(\tilde{m}_{1}, \dots, \tilde{m}_{T})),
% \end{equation}
% \end{small}
\vspace{-4mm}
\begin{small}
\begin{equation}
    \mathcal{L}_{act} = \mathrm{CE}\left(p^a, \mathrm{SM}\left(\mathrm{MLP}(\mathrm{Avg}(\tilde{m}_{1}, \dots, \tilde{m}_{T}))\right)\right),
\end{equation}
%\vspace{-2mm}
\end{small}where $p^a$ stands for the ground-truth activity category,  $\mathrm{SM}$ denotes the softmax operation, and $\mathrm{CE}$ denotes the cross-entropy loss function.

(2) \textit{Spatial Relation Detection.} Accurately distinguishing spatial relations between human and scene context is crucial for modeling interactive cues. To enhance this capability, we introduce a \textit{spatial relation detection} task to classify human-scene spatial relations. Specifically, we define 8  categories (\eg, `facing') to characterize human-object spatial relations. Given the scene embedding $s_i$ and motion embedding $m_t$, AInt module predicts the spatial relation between the $i$-th object and human motion at $t$-th frame, supervised by a cross-entropy loss:

%  \begin{small}
% \begin{equation}
%     \mathcal{L}_{spa} = \sum_{i,t}  \mathrm{CE}\left(p^s_{it},\hat{p}^s_{it}\right), \ \hat{p}^s_{it} = W^{spa}_{s}(s_i) \cdot W^{spa}_{m}(m_t),
% \end{equation}
%  \end{small}
    \vspace{-3mm}
 \begin{small}
\begin{equation}
    \mathcal{L}_{spa} = \sum_{i,t}  \mathrm{CE}\left(p^s_{it}, \mathrm{SM}\left(W^{spa}_{s}(s_i) \cdot W^{spa}_{m}(m_t)\right)\right),
\end{equation}
 \end{small}where $p^s_{it}$ stands for the ground-truth spatial relation label between the $i$-th object and $t$-th motion frame, $W^{spa}_{s}$ and $W^{spa}_{m}$ are linear projection weights.

(3) \textit{Contact Detection.} Another crucial aspect for human-scene interactions is physical contact between human body and surrounding objects. To capture these cues, we introduce a \textit{contact detection} task, which predicts whether an object is in contact with a specific human body part, supervised by a binary cross-entropy loss:
% \begin{equation}
%     \mathcal{L}_{cont} = \mathrm{BCE}(p^c_{it}, \hat{p}^c_{it}), \
%     \hat{p}^c_{it} = W^{cont}_{s}(s_i) \cdot W^{cont}_{m}(m_t),
% \end{equation}

\vspace{-2mm}
\begin{small}
\begin{equation}
    \mathcal{L}_{cont} = \sum_{i,t} \mathrm{BCE}(p^c_{it}, \sigma(W^{cont}_{s}(s_i) \cdot W^{cont}_{m}(m_t))),
\end{equation}
\end{small}where $p^c_{it}$ represents the ground-truth contact label, with $\left[p^c_{it}\right]_l=1$ (or $0$) indicating that the $i$-th object is in contact (or not) with the  $l$-th body joint at $t$-th motion frame, $W^{cont}_{s}$ and $W^{cont}_{m}$ are projecting weights.  $\sigma$ denotes the sigmoid function  and $\mathrm{BCE}$ denotes binary cross-entropy function.

\noindent\textbf{Layout-trajectory Position Encoding (LTP) Module.} \ Traditional position encoding in MLLMs primarily model sequential relationships among tokens, overlooking the complex spatiotemporal relationships between human and their surrounding environment. To this end, we propose LTP, which generates position embeddings based on spatial locations and temporal orders of human and scene input. By globally aligning spatial and temporal information across human motion and scene modalities, LTP enhances contextural awareness, enabling each modality to more effectively incorporate relevant information from the other.

As shown in~\cref{fig:his-gpt},  LTP module consists of a Spatial Fourier-transform (SF) and a Temporal Fourier-transform (TF) layer to encode 3D spatial coordinates and temporal information, respectively. Specifically, given a 3D coordinate $\mu=\left[x,y,z\right]$ and a timestamp  $t \in \left[1,T\right]$, ST and TF layers are implemented as follows:

\vspace{-3mm}
\begin{small}
\begin{equation}
    SF(\mu)=\mathrm{sincos}(\phi_{SF}\cdot 2\pi\mu), TF(t)=\mathrm{sincos}(\phi_{TF}\cdot 2\pi t),
\end{equation}
\end{small}where $\phi_{SF}$ and $\phi_{TF}$ are linear projection weights, and $\mathrm{sincos(\cdot)}$ denotes the concatenation of sine and cosine results along latent dimension. 

Leveraging SF and TF layers,  for human motion modality, LTP generates a position encoding vector $e_t^m=SF(\mu_t) + TF(t)$ for the $t$-th motion frame, based on its 3D location $\mu_t=\left[x_t^m,y_t^m,z_t^m\right]$ and timestamp $t$.
For 3D scene modality, LTP module yields a position encoding vector $e_i^s = SF\left(\mu_i\right) + \frac{1}{T} \sum_{t} TF\left(t\right)$ for the $i$-th object, based on its 3D location $\mu_i=[x_i^s,y_i^s,z_i^s]$.
Note that we apply averaging to the temporal fourier transformations across all motion timestamps, as the object presents throughout the entire motion sequence.
Finally, we aggregate the position encodings into the embeddings of each modality as: $f_i^s = s_i + e_i^s$, $f_t^m = m_t + e_t^m$.
In this manner, we obtain latent features $F^s=\left\{ f_i^s \right\}_{i=1}^N$ and $F^m=\left\{ f_t^m \right\}_{t=1}^T$ for scene and motion modality, respectively. %These features incorporate modality-shared position information, thereby enhancing the model’s ability to capture human-scene interactions.

\noindent\textbf{LLM.} \ After the LTP module, the latent scene feature $F^s$ and motion feature $F^m$ are fed into a decoder-only LLM.  Given the test instruction $\mathcal{I}$ and answer $\mathcal{A}$,   the LLM predicts the probability distribution of potential next answer token at each step, $P\left(\mathcal{A}_{\left[n\right]} | F^{s}, F^{m}, \mathcal{I}, \mathcal{A}_{\left[<n\right]}\right)$,  in an autoregressive manner. The objective is to maximize the log-likelihood of this predicted probability distribution, denoted as $\mathcal{L}_{llm} = - \sum_{n} \log 
        P\left(\mathcal{A}_{\left[n\right]} | F^{s}, F^{m}, \mathcal{I}, \mathcal{A}_{\left[<n\right]}\right)$.
% \begin{equation}
%     \mathcal{L}_{llm} = - \sum_{n} \log 
%         P_{\theta}\left(a_{\left[n\right]} | F^{s};F^{m};\mathcal{I}; a_{\left[<n\right]}\right),
% \end{equation}
% where  $\theta$ represents the trainable parameters. 

\input{tables/his-bench}

\subsection{Training}

To effectively align the 3D scene and human modalities with the LLM, we propose a two-stage training strategy: 

\noindent
\textbf{Stage1: Modality alignment}: In this stage, we use the annotation pipeline described in~\cref{sec:his-bench_data generation} to craft detailed HIS captions for aligning input modalities with LLM. Additionally, we add scene captions and motion captions to further enhance the alignment. This stage uses the autoregressive loss of LLM along with the auxiliary tasks in AInt module for training: $\mathcal{L}=\mathcal{L}_{llm} + \lambda_{act}\mathcal{L}_{act} + \lambda_{spa}\mathcal{L}_{spa} + \lambda_{cont}\mathcal{L}_{cont}$,
where $\lambda_{act}$, $\lambda_{spa}$ and $\lambda_{cont}$ are hyperparameters. 

\noindent
\textbf{Stage2: HIS instruction tuning}: In this stage, we synthesize a diverse instruction-following HIS data corpus, which covers a wide range of capabilities and formats for tuning. We only fine-tune HIS-GPT with $\mathcal{L}_{llm}$ to ensure the quality of instruction following.

In total, our training data comprises 60k visual captions and 700k instruction tuning samples, covering over 750 diverse scenes. More details about the training data are provided in Appendix~\ref{sec:training_data}.

%% file: tables/his-bench.tex
\begin{table*}
\centering
    \caption{Quantitative evaluation results on HIS-Bench. We run the evaluation for three times and report the average score for each dimension. The full score for each dimension is 100. \textbf{`Avg.'} is the average score across all 16 dimensions. The best and second-best results are \textbf{boldfaced} and \underline{underlined}, respectively.}
    \vspace{-2mm}
\begin{adjustbox}{width=\linewidth,center}
\tablestyle{6pt}{1.05}
\begin{tabular}{l|cc|ccc|ccc|cc|cc|c|ccc|c}
\toprule
     \multirow{2}{*}{\textbf{Methods}} & \multicolumn{2}{c|}{\textbf{Activity}} & \multicolumn{3}{c|}{\textbf{Spatial Relationship}} & \multicolumn{3}{c|}{\textbf{Human-object Interaction}} & \multicolumn{2}{c|}{\textbf{Analysis}} & \multicolumn{2}{c|}{\textbf{Prediction}} & \multirow{2}{*}{\textbf{Dialogue}} & \multicolumn{3}{c|}{\textbf{Planning}} & \multirow{2}{*}{\textbf{Avg.}} \\
     & AC & SA & HP & BO & OO & IT & IO & CP & FA & SA & IP & MP & & HT & LT & NA & \\
\midrule
    \multicolumn{18}{l}{\textbf{\textit{3D Scene MLLMs}}} \\
\midrule
    LL3DA~\cite{chen2024ll3da} & 9.0 & 4.0 & 3.5 & 4.7 & 19.0 & 4.0 & 10.5 & 11.7 & 6.5 & 17.2 & 4.2 & 6.3 & 4.7 & 1.0 & 0.3 & 0.0 & 6.7 \\
    Chat-Scene~\cite{huang2024chat} & 1.8 & 16.5 & 0.5 & 6.5 & 5.2 & 3.0 & 24.3 & 14.7 & 3.7 & 18.3 & 6.3 & 7.3 & 3.5 & 10.0 & 8.8 & 1.3 & 8.2 \\
\midrule
    \multicolumn{18}{l}{\textbf{\textit{Vision LLMs}}} \\
\midrule
    GPT-4v~\cite{2023GPT4VisionSC} & 10.5 & 22.3 & 7.2 & 34.7 & 25.0 & 24.2 & 49.2 & 24.7 & 5.7 & 28.3 & 12.2 & 16.0 & \textbf{58.7} & 33.5 & 24.2 & 10.5 & 24.2 \\
    GPT-4o~\cite{hurst2024gpt} & 24.3 & \underline{36.0} & \underline{9.7} & \underline{36.5} & \underline{31.3} & \underline{32.7} & 46.0 & 31.2 & \underline{31.3} & \underline{39.7} & 23.3 & \underline{17.7} & 36.5 & \underline{54.3} & \underline{35.3} & \underline{15.0} & \underline{31.3} \\
    Qwen-VL-max~\cite{bai2023qwen} & \underline{25.3} & 32.0 & 7.7 & 31.8 & 13.2 & 25.0 & \underline{54.7} & \underline{31.7} & 9.0 & 17.8 & 19.3 & 9.7 & 33.0 & 31.5 & 26.2 & 8.7 & 23.5 \\
    Qwen2.5-VL~\cite{bai2025qwen2} & 10.2 & 11.0 & 5.5 & 27.3 & 18.3 & 16.7 & 49.0 & 29.7 & 2.8 & 20.3 & 12.5 & 16.7 & 15.5 & 21.5 & 20.0 & 7.7 & 17.8 \\
    LLaVA-OV~\cite{li2024llava} & 15.3 & 7.7 & 9.2 & 16.0 & 14.3 & 16.7 & 41.3 & 27.7 & 1.0 & 14.5 & 9.5 & 7.5 & 16.7 & 17.8 & 8.2 & 4.0 & 14.2 \\
    LLaVA-Video~\cite{zhang2024video} & 11.3 & 16.2 & 4.0 & 20.8 & 9.0 & 17.8 & 27.8 & 29.0 & 13.8 & 21.5 & 12.5 & 14.0 & 20.8 & 19.7 & 16.0 & 6.2 & 16.3 \\
\midrule
    \multicolumn{18}{l}{\textbf{\textit{LLMs w/ Frame Captions}}} \\
\midrule
    LLaVA-OV~\cite{li2024llava}+GPT-4~\cite{achiam2023gpt} & 9.0 & 10.3 & 5.5 & 22.3 & 16.0 & 14.7 & 29.3 & 18.0 & 2.7 & 21.2 & \underline{27.5} & 13.0 & 53.5 & 22.7 & 15.3 & 5.5 & 17.9 \\
    Qwen-VL-max~\cite{bai2023qwen}+GPT-4~\cite{achiam2023gpt} & 5.3 & 6.0 & 3.2 & 8.3 & 10.0 & 3.5 & 29.7 & 13.0 & 0.6 & 6.0 & 14.5 & 5.3 & 22.0 & 6.5 & 1.7 & 4.8 & 8.8 \\
\midrule
    \multicolumn{18}{l}{\textbf{\textit{LLMs w/ Scene\&Motion Captions}}} \\
\midrule
    LL3DA~\cite{chen2024ll3da}+AvatarGPT~\cite{zhou2024avatargpt}+GPT-4~\cite{achiam2023gpt} & 1.3 & 0.5 & 2.5 & 5.7 & 0.3 & 2.5 & 21.5 & 12.8 & 0.0 & 3.7 & 6.0 & 2.7 & 13.3 & 3.3 & 2.7 & 1.0 & 5.0 \\
\midrule
    \multicolumn{18}{l}{\textbf{\textit{HIS Foundation Models} \ (Ours)}} \\
\midrule
    \textbf{HIS-GPT} & \textbf{39.3} & \textbf{49.8} & \textbf{37.0} & \textbf{57.3} & \textbf{32.0} & \textbf{52.8} & \textbf{58.3} & \textbf{55.5} & \textbf{33.8} & \textbf{48.2} & \textbf{50.5} & \textbf{50.0} & \underline{53.2} & \textbf{55.7} & \textbf{58.0} & \textbf{48.0} & \textbf{48.7} \\
\bottomrule
\end{tabular}
\end{adjustbox}
\label{tab:his-bench}
\vspace{-5mm}
\end{table*}

%% file: sec/5_experiments.tex
\section{Experiments}
\label{sec:experiments}

\input{tables/ablation-component}

\subsection{Experimental Setup}

\textbf{HIS-QA Baselines.}  \ 
Inspired by the recent advances in vision-language models, we investigate how well these models could address the proposed HIS-QA task.
\textbf{(1) 3D scene LLMs.}  Current 3D scene LLMs are incapable of processing sequential human motion. To adapt these models for HIS-QA, we convert the human body from a randomly selected frame into a point cloud format, and input it alongside the scene mesh into the 3D scene LLM. We employ  LL3DA~\cite{chen2024ll3da} and Chat-Scene~\cite{huang2024chat} for evaluation.
 \textbf{ (2) Vision LLMs.}  Since existing vision LLMs cannot directly process 3D input,  we render HIS data into video segments and input them into vision LLMs. We select models  from the GPT~\cite{hurst2024gpt}, Qwen~\cite{bai2023qwen}, and LLaVA~\cite{li2024llava} families. 
\textbf{ (3) LLMs w/ Frame Captions.}  To leverage strong image captioners, we first derive frame-level captions from rendered HIS videos, and input these captions into a LLM to answer HIS questions. We adopt Qwen-vl-max~\cite{bai2023qwen} and LLaVA-OV~\cite{li2024llava} as captioners, and GPT-4~\cite{achiam2023gpt} as LLM. 
\textbf{ (4) LLMs w/ Scene\&Motion Captions.} To extract linguistic information from HIS data, we separately use captioners for 3D scene and 3D human motions, and feed these scene and motion captions into an LLM to perform HIS tasks. Specifically, we use LL3DA~\cite{chen2024ll3da}, AvatarGPT~\cite{zhou2024avatargpt}, and GPT-4~\cite{achiam2023gpt} as scene captioner, motion captioner, and LLM respectively.  The detailed implementation of HIS-QA baselines is provided in Appendix~\ref{sec:details_HIS_QA_baselines}.

\noindent
\textbf{Implementation Details for HIS-GPT.} \
We adopt Vicuna-1.5~\cite{chiang2023vicuna} as LLM backbone, and AdamW~\cite{loshchilov2017decoupled} optimizer for training. HIS-GPT is trained in two stages: stage 1 runs for $100$k steps with a learning rate of $1 \times 10^{-4}$, while stage 2 runs for $50k$ steps with a reduced learning rate of $5 \times 10^{-5}$. The batch size is set to $16$ for both stages. To preserve the original capabilities of the backbones, we keep the scene encoder, motion encoder and LLM frozen throughout training,  fine-tuning only the projection layers, AInt and LTP modules. The loss weights $\lambda_{act}$, $\lambda_{spa}$ and $\lambda_{cont}$ are set to $0.5$, $0.5$ and $0.1$, determined by grid search. %Please refer to Appendix D.1 for more details.

\noindent \textbf{Evaluation Metrics of HIS-Bench.} \
Considering that HIS-Bench consists of open-ended questions, we use GPT-4 as an automatic evaluator to assess answer correctness. Following~\cite{cheng2024egothink}, we prompt GPT-4 to assign a score between 0 and 2 for each answer. Since each task in HIS-Bench consists of 50 questions, the full score for each task is 100.

\subsection{Quantitative Results}
\cref{tab:his-bench} provides the quantitative results on HIS-Bench. Based on the results, we summarize our findings as follows:

\input{tables/ablation-tuning}

\noindent
\textbf{Question types.} \ From~\cref{tab:his-bench}, we observe that almost all models perform relatively well on dialogue (SD) and task-planning (HT, LT), likely because these dimensions are closely aligned with the conversation and reasoning abilities inherent in original LLMs. In contrast, tasks requiring a strong understanding of spatial characteristics, such as Human Position (HP) and Navigation (NA), present significant challenges, highlighting the need for the development of more advanced spatial interaction modeling capabilities.

\begin{figure*}
\vspace{-3mm}
\centering
    \includegraphics[width=0.75\linewidth]{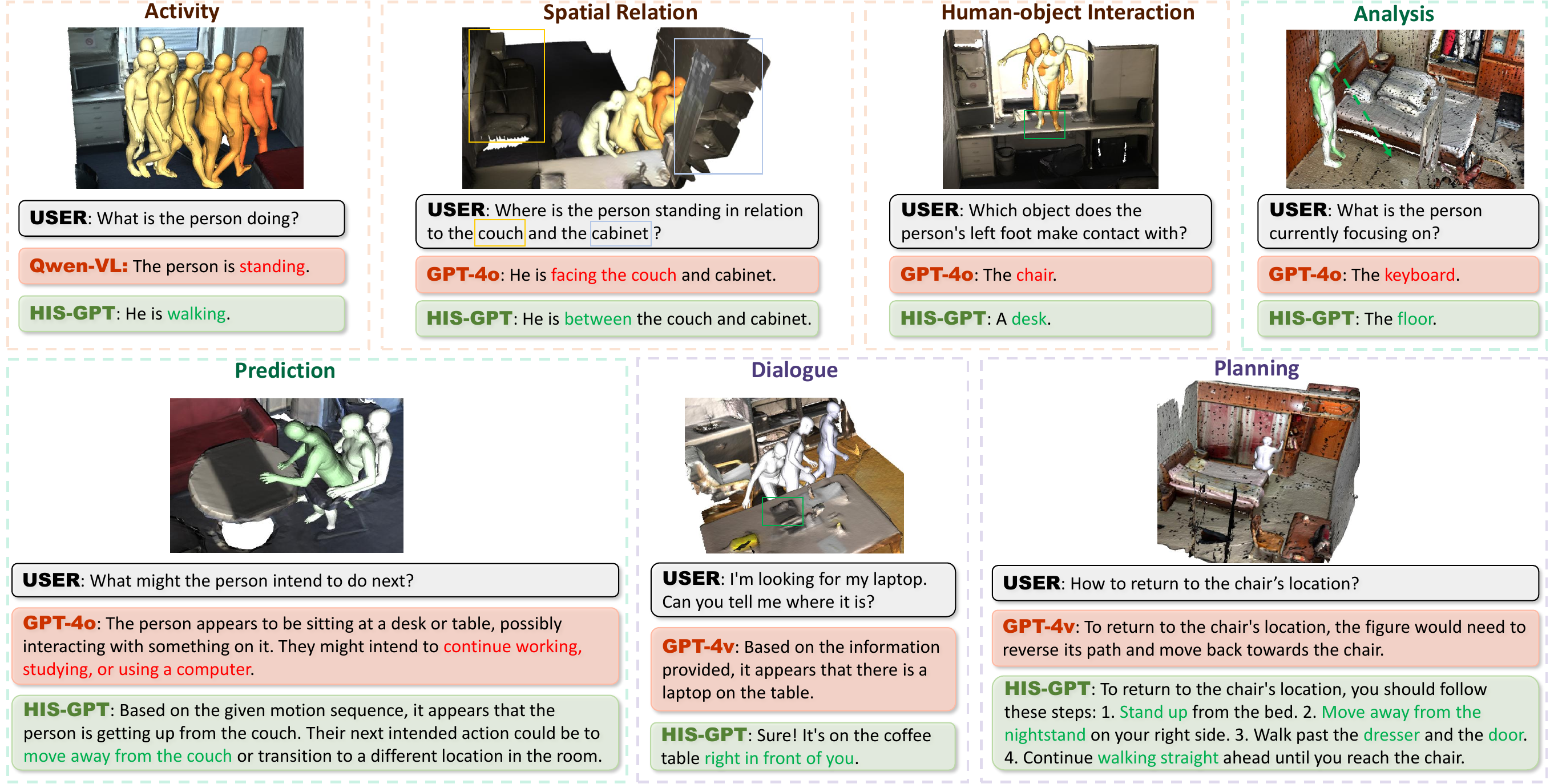}
    \caption{Qualitative comparisons of HIS-GPT and other baselines on HIS-QA. Red/green color denotes wrong/correct outputs.}
    \label{fig:qa examples}
    \vspace{-3mm}
\end{figure*}

\noindent
\textbf{HIS-QA baselines results.} \ Among all the baselines, vision LLMs demonstrate significantly superior performance compared to other model families. This can be attributed to their outstanding instruction-following abilities which enable them to effectively follow HIS-QA questions, as well as their robust visual generalization to rendered videos. While 3D scene LLMs excel in 3D understanding, their ability to generalize to 3D human input is limited due to the absence of human-centric data in their training corpus. Similarly, LLMs w/ captions are restricted by the lack of detailed spatial information and human-scene interactions within their generated captions, leading to a weaker understanding of complex human-in-scene behaviors. 

\noindent
\textbf{HIS-GPT results.} \ As evident in~\cref{tab:his-bench}, HIS-GPT significantly outperforms all HIS-QA baselines, achieving an average score exceeding the highest-performing baseline, GPT-4o, by $17.4$ points. Compared to other vision-language models, HIS-GPT demonstrates particular strength in tasks requiring a nuanced understanding of spatial relations between humans and their 3D surroundings, such as Human Position (HP) and Contact Part (CP). Also, HIS-GPT performs well in prediction tasks, showcasing its ability to accurately infer human states and perform complex reasoning.

\subsection{Ablation Studies}
We conduct ablation studies to validate the effectiveness of HIS-GPT. Additional ablations,  including loss weight and LLM tuning strategy, are provided in Appendix~\ref{sec:supp-quantitative}. 

\noindent\textbf{Ablations on AInt module.} \ \cref{tab:ablation-components} reports the ablation studies on the AInt module.  The results indicate that integrating AInt increases the average score on HIS-Bench by $1.1$, demonstrating its effectiveness for human-in-scene understanding. 
To further analyze its impact, we break down the contributions of individual sub-tasks within the AInt module. As shown in \cref{tab:ablation-components}, activity classification (act), spatial relation detection (spa), and contact detection (cont) tasks improve their corresponding HIS-Bench core tasks  (Activity, Spatial Relationship, and HoI) by $1.3$, $0.9$, and $1.7$, respectively.  These results indicate that explicitly modeling fine-grained human-scene interactions through AInt substantially benefits the overall capabilities of HIS-GPT.

\noindent\textbf{Ablations on LTP module.} \ 
As shown in \cref{tab:ablation-components}, integrating the LTP module leads to a $3.0$ average score gain on HIS-Bench, demonstrating its effectiveness. Furthermore, when AInt and LTP are used jointly, they achieve a significant $5.7$ performance gain over the baseline. This result highlights the complementary nature of these modules, suggesting that combining fine-grained human-scene interaction modeling with structured spatial-temporal encoding can further enhance the model's ability to comprehensively understand human activities in 3D environments.

\noindent\textbf{Ablations on Training Strategy.} \ \cref{tab:ablation-tuning} presents ablation study on the two-stage training strategy. The results indicate that both modality alignment (Stage 1) and instruction tuning (Stage 2) are essential for effectively training HIS-GPT. Additionally, incorporating scene and motion caption data in Stage 1 leads to a rise of $2.9$ in average score, validating their effectiveness in facilitating modality alignment.

% \noindent\textbf{Hyper-parameter Analysis on Loss Weight.} \ To determine the optimal loss weights in training stage 1, we conduct grid search on the weight parameters. Results validate that the best performance is achieved at our hyper-parameter setting: $\lambda_{act}=0.5, \lambda_{spa}=0.5$ and $\lambda_{cont}=0.1$. Detailed results are shown in Appendix.

\subsection{Qualitative Results}
\cref{fig:qa examples} presents qualitative examples of HIS-GPT across various HIS-QA tasks. Compared to baseline models, HIS-GPT gives more accurate answers in basic perceptions about human activities, spatial relation to scene, and interaction with objects. Moreover, HIS-GPT generates highly plausible responses in reasoning and prediction tasks, showcasing a strong understanding of human behavior within the scene. Additionally, HIS-GPT excels in dialogue and planning tasks, which are crucial for embodied AI applications. Notably, while GPT-4v frequently produces generic or uninformative responses that are not helpful enough for users to address their problems, HIS-GPT provides constructive replies with situated knowledge (\eg, `right in front of you') and detailed guidance (\eg, `stand up', `walk straight') that can effectively assist users in real-world scenarios.
More qualitative results are provided in Appendix~\cref{sec:supp-qualitative}.

%% file: tables/ablation-component.tex
\begin{table}[]
\centering
    \caption{Ablations on the key components of HIS-GPT. `act', `spa' and `cont' denotes the activity classification, spatial relation detection and human-scene contact detection task in AInt module. `PE' denotes position encoding methods.}
    \vspace{-2mm}
\begin{adjustbox}{width=\linewidth,center}
\tablestyle{7pt}{1.0}
\begin{tabular}{c|ccc|c|c|ccc}
\toprule
    \multirow{2}{*}{\textbf{Methods}} & \multicolumn{3}{c|}{\textbf{AInt}} & \multirow{2}{*}{\textbf{PE}} & \multicolumn{4}{c}{\textbf{HIS-Bench}} \\
    & act & spa & cont && Act. & Spa. & HoI. & \textbf{Avg.} \\
\midrule
    1 & & & & sine & 41.8 & 34.7 & 45.8 & 43.0   \\
\midrule
    2 & \Checkmark & \Checkmark & \Checkmark & sine & 43.5 & 35.3 & 51.0 & 44.1   \\
    3 & & & & LTP & 43.5 & 38.8 & 50.3 & 46.0   \\
    4 & \Checkmark & & & LTP & \textbf{44.8} & 36.5 & 47.5 & 45.3   \\
    5 & & \Checkmark & & LTP & 42.4 & 39.7 & 48.8 & 47.3   \\
    6 & & & \Checkmark & LTP & 43.3 & 38.5 & 52.0 & 46.9   \\
\midrule
    7(Ours) & \Checkmark & \Checkmark & \Checkmark & LTP & 44.6 & \textbf{42.1} & \textbf{55.5} & \textbf{48.7}   \\
\bottomrule
    
\end{tabular}
\end{adjustbox}
\label{tab:ablation-components}
\vspace{-4mm}
\end{table}

%% file: tables/ablation-tuning.tex
\begin{table}[]
\centering
    \caption{Ablations on the training strategy of HIS-GPT. `HIS', `Scene' and `Motion' denotes the usage of HIS, scene and motion data in stage 1 training.}
    \vspace{-2mm}
\begin{adjustbox}{width=\linewidth,center}
\tablestyle{3pt}{1.0}
\begin{tabular}{ccc|c|ccccccc|c}
\toprule
    \multicolumn{3}{c|}{\textbf{Stage 1}} & \multirow{2}{*}{\textbf{Stage 2}} & \multicolumn{8}{c}{\textbf{HIS-Bench}} \\
    HIS & Scene & Motion && Act. & Spa. & HoI. & Ana. & Pre. & Dia. & Pla. & \textbf{Avg.} \\
\midrule
    \Checkmark & \Checkmark & \Checkmark & & 39.3 & 30.2 & 41.0 & 32.8 & 40.5 & 35.5 & 41.8 & 37.5 \\
    & & & \Checkmark & 39.0 & 31.3 & 47.8 & 37.0 & 46.0 & 47.5 & 50.8 & 42.6 \\
    \Checkmark & & & \Checkmark & 42.2 & 36.5 & 51.8 & 37.8 & 47.0 & 41.5 & 52.3 & 45.8 \\
    \Checkmark & \Checkmark & & \Checkmark & 39.0 & 39.7 & 46.9 & 38.8 & 46.0 & 42.5 & 51.2 & 44.0 \\
    \Checkmark & & \Checkmark & \Checkmark & 39.5 & \textbf{42.8} & 49.7 & \textbf{41.0} & 46.0 & 48.5 & 52.5 & 46.0 \\
\midrule
    \Checkmark & \Checkmark & \Checkmark & \Checkmark & \textbf{44.6} & 42.1 & \textbf{55.5} & \textbf{41.0} & \textbf{50.3} & \textbf{53.2} & \textbf{53.9} & \textbf{48.7} \\
\bottomrule
\end{tabular}
\end{adjustbox}
\label{tab:ablation-tuning}
\vspace{-5mm}
\end{table}

%% file: sec/X_suppl.tex
% \clearpage
% \setcounter{page}{1}
% \renewcommand{\thefigure}{S\arabic{figure}}
% \setcounter{figure}{0}
% \renewcommand{\thetable}{S\arabic{table}}
% \setcounter{table}{0}
% \renewcommand\thesection{\Alph{section}}
% \setcounter{equation}{0}
% \renewcommand\theequation{A.\arabic{equation}}
% \maketitlesupplementary
\appendix

% In Appendix \ref{sec:supp-hisbench details}, we provide a comprehensive overview of the implementation details of HIS-Bench. Appendix \ref{sec:supp-experiment details} elaborates on the experimental setup, including the  HIS-QA baselines and our proposed HIS-GPT model. In Appendix \ref{sec:supp-hisbench samples}, we present additional data samples of constructed HIS-Bench. Appendix \ref{sec:supp-quantitative} and \ref{sec:supp-qualitative} provide supplementary quantitative and qualitative results, offering deeper insights into the performance and limitations of HIS-GPT. 

\section{HIS-Bench Details}
\label{sec:supp-hisbench details}

In this part, we present additional details of HIS-Bench, including a detailed description of the multi-faceted text annotation process (Appendix \ref{sec:details_on_multi}), the prompts used for question-answer generation (Appendix \ref{sec:prompts_for_QA}), the web interface designed for human annotation  (Appendix  \ref{sec:interface_for_human}), user studies on benchmark quality (Appendix  \ref{sec:user_study}), and studies on the accuracy of using GPT as automatic evaluator (Appendix  \ref{sec:gpt_evaluation_acc}).

\subsection{Details on Multi-Faceted Text Annotation}
\label{sec:details_on_multi}

In Sec~\ref{sec:his-bench_data generation}, we present the multi-faceted text annotation process employed to generate rich linguistic labels for HIS datasets. In this part, we further elaborate on the implementation details, including scene annotation and frame-level contact/position annotation.

\noindent\textbf{Scene Annotation.} \ Following SceneVerse~\cite{jia2024sceneverse}, we first utilize a 3D scene segmentor~\cite{schult2023mask3d} to obtain 6D bounding boxes and semantic labels of each object in the scene. Next, we construct a scene graph, where nodes represent objects, and edges capture the spatial relationships between two objects. 
From this scene graph, we extract triplets consisting of two objects and their relationship such as (sofa, near, chair). Finally, we apply predefined natural language templates to transform these triplets into referring expressions for each object, \eg, \textit{`The sofa is near the chair'}. For details on the definitions of spatial relationships and the templates, please refer to~\cite{jia2024sceneverse}.

\noindent\textbf{Frame-level Contact Annotation.} \ For the annotation of frame-level human contact with the scene, we represent the annotations as tuples \textit{(body joint, anchor object)}, indicating that a specific body joint is in contact with an anchor object in the scene. In details, we define 22 body joints based on the 3D human representations following~\cite{wang2022humanise}. To determine contact, we compute the closest distance between each body joint and the point cloud of each object,  labeling them as in contact if the distance is below a threshold $\epsilon$, which we set to $0.1$ in practice. The full list of body joints is provided below:

\vspace{2mm}
\noindent\fbox{
\begin{minipage}{\dimexpr0.95\linewidth}
    pelvis, left hip, right hip, lower spine, left knee, right knee, middle spine, left ankle, right ankle, upper spine, left foot, right foot, neck, left collar, right collar, head, left shoulder, right shoulder, left elbow, right elbow, left wrist, right wrist.
\end{minipage}
}
\vspace{2mm}

\begin{figure}
    \centering
    \includegraphics[width=0.95\linewidth]{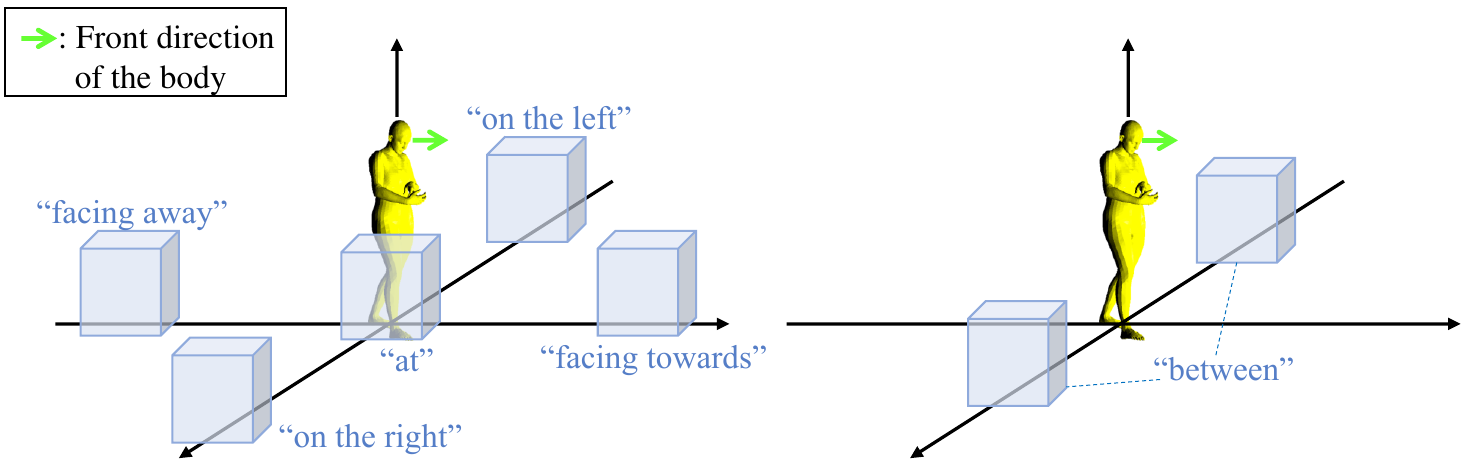}
    \caption{Illustrations of the definition of human-object orientations. We define six types of orientations: `facing towards', `on the left', `on the right', `facing away', `at', and `between'.}
    \label{fig:orientation definition}
\end{figure}

\noindent\textbf{Frame-level Position Annotation.} \ For frame-level position annotation  within the scene, we represent the annotation as triplets \textit{(orientation, distance, anchor object)}. Each triplet captures the relative orientation and distance of an anchor object with respect to the human body,  providing a structured representation of the human’s spatial context. 
Specifically, \textit{orientation} is defined as the relative direction of the anchor object with respect to the human. We categorize orientation into six types: `facing towards', `on the left', `on the right', `facing away', `at' and `between', based on the angle between the object's direction and the human's facing direction, as illustrated  in~\cref{fig:orientation definition}. Notably, the `between' category describes a scenario where one object is positioned to the left and another object to the right of the person.
\textit{Distance} is defined as the horizontal distance (on xy-plane) between the center point of the anchor object and the pelvis point of human body.

\subsection{Prompts for Question-Answer Generation}
\label{sec:prompts_for_QA}

\begin{figure*}
    \centering
    \includegraphics[width=0.95\linewidth]{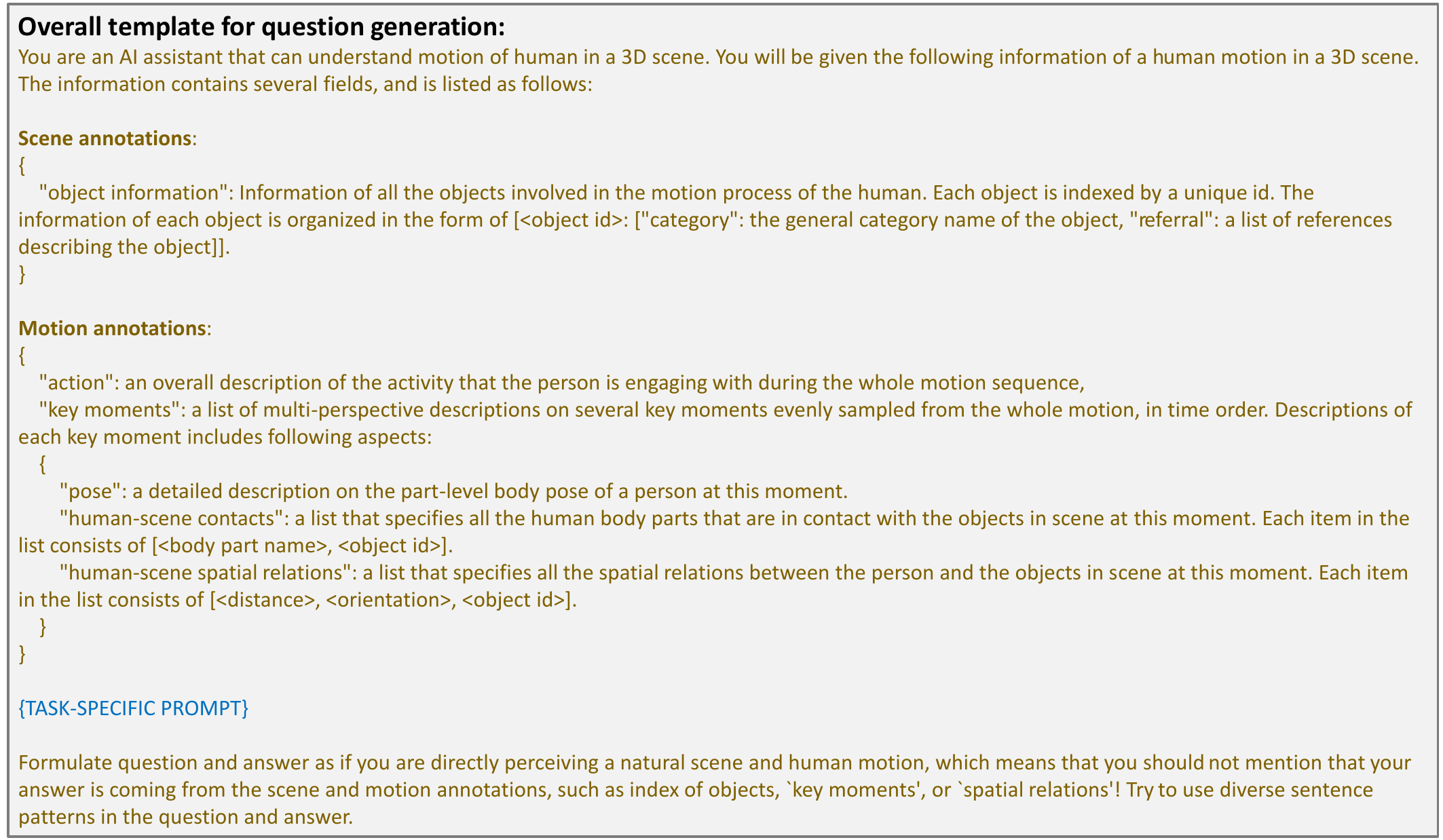}
    \caption{The overall prompt template for HIS-Bench question-answer generation. The `\{TASK-SPECIFIC PROMPT\}' placeholder is filled with prompts tailored to specific HIS-QA tasks.}
    \label{fig:overall_prompt}
\end{figure*}

\begin{figure*}
    \centering
    \includegraphics[width=0.95\linewidth]{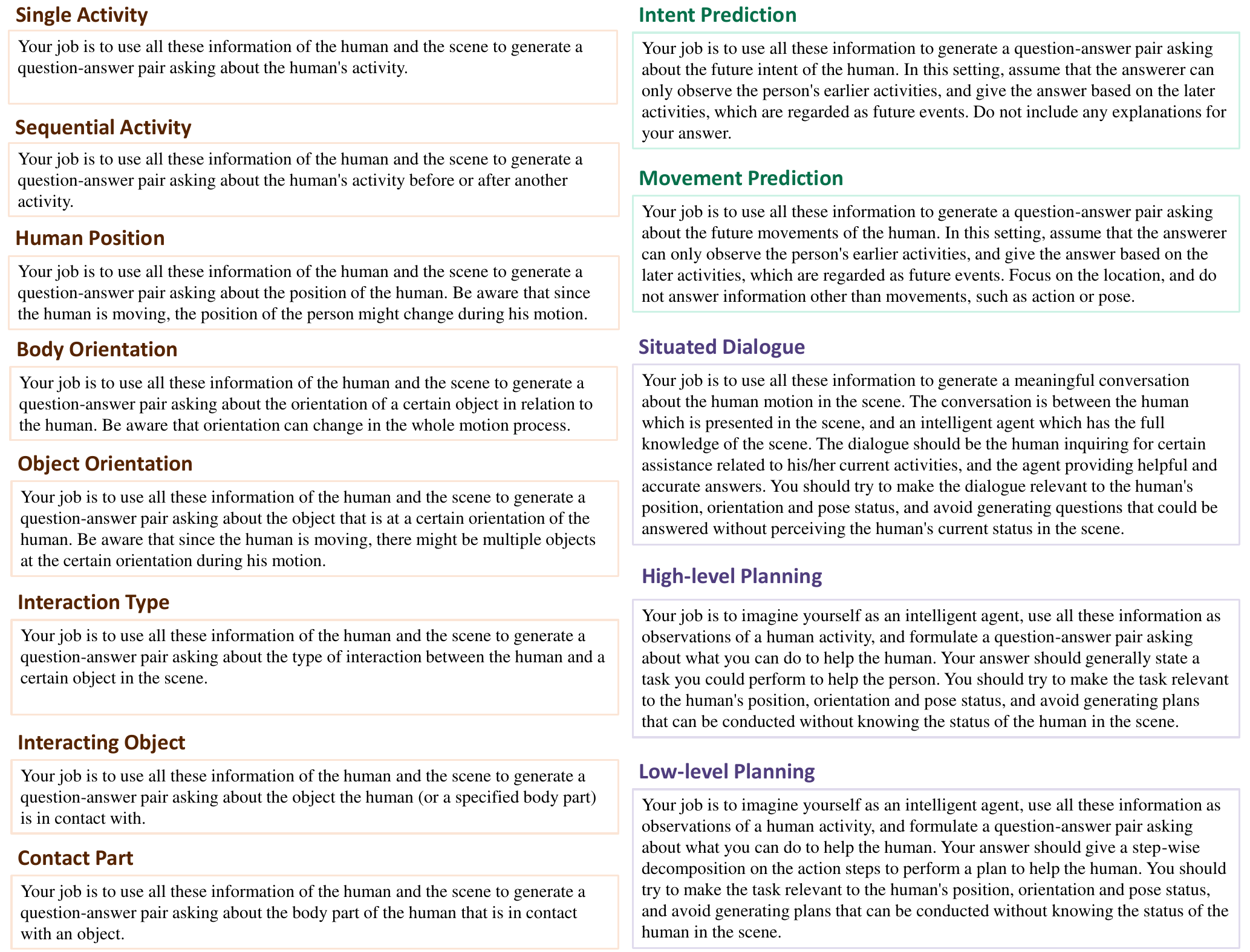}
    \caption{The task-specific prompts for each HIS-QA task.}
    \label{fig:task-specific_prompt}
\end{figure*}

We formulate a large proportion of HIS-Bench questions by prompting GPT-4~\cite{achiam2023gpt}. In~\cref{fig:overall_prompt} and~\cref{fig:task-specific_prompt}, we present all prompts designed for the GPT-assisted generation process. Specifically, we create a general template for generating all questions, and then fill it with task-specific instructions when generating questions for each sub-task.

\subsection{Interface for Human Annotation}
\label{sec:interface_for_human}

For three of the 16 sub-tasks in HIS-Bench, namely \textit{focus analysis}, \textit{situated analysis} and \textit{navigation}, we employ human annotators to write question-answer pairs based on the raw HIS data. To facilitate this process,  we design a web interface for annotation. On the interface, annotators are shown the rendered video of the HIS data sequence, and are required to input the question, answer, start timestamp, and end timestamp to create a data sample. The interface also provides detailed instructions on how to annotate, along with examples of good and bad annotations.

\input{tables/supp-userstudy}

\subsection{User Study of Benchmark Quality}
\label{sec:user_study}

Considering that the majority of QA pairs in HIS-Bench are generated in an automatic manner, we carry out a user study to verify the quality of these QA pairs. Specifically, we employ three human annotators and let them grade every QA sample in HIS-Bench from the following four aspects: question answerability, question clarity, answer correctness and question difficulty. For each aspect, annotators are instructed to give an integer score from 1 to 5 (1 is lowest, and 5 is highest). Then, for a more objective reference, we additionally ask the annotators to grade 100 QA samples from another human-annotated 3D scene benchmark, OpenEQA~\cite{majumdar2024openeqa}, and compare the average grades between OpenEQA and HIS-Bench. As shown in~\cref{tab:user_study}, in all evaluated aspects, HIS-Bench achieves similar scores with the human-annotated and verified OpenEQA benchmark, proving the quality of the automatically synchronized QA pairs in HIS-Bench.

\subsection{Accuracy of GPT Evaluation}
\label{sec:gpt_evaluation_acc}

As HIS-Bench consists of open-ended questions, we deploy GPT-4~\cite{achiam2023gpt} as the evaluation tool for the answers. Here we conduct a detailed analysis of the accuracy of GPT evaluations. We derive the model-generated (here we use GPT-4o~\cite{hurst2024gpt}) answers on HIS-Bench, comparing the consistency of GPT evaluations and human judgements on these answers. Three human evaluators are invited to grade the answers from aspects including accuracy, information completeness, logical soundness and grammar correctness, by giving each dimension a score from 1 to 5. Then, a human-evaluated score for each answer is conducted by averaging the scores of each dimension. Using the scores, following~\cite{cheng2024egothink}, we calculate the Pearson correlation score ($\in [-1,1]$, $> 0$ means positive correlation) between GPT and human evaluations. Results show that the Pearson correlation is $0.54$, reflecting that GPT evaluations are accurate since they are highly consistent with humans.

Moreover, we verify the reproducibility of using LLM as evaluators by adopting an open-sourced LLM, Qwen2.5-7B~\cite{bai2025qwen2} as the evaluator. We find that the Pearson correlation score between Qwen2.5-7B and GPT-4 is $0.75$, and their judgement scores on the answers of multiple models are consistent, as shown in~\cref{tab:evaluator_reproduce}, the evaluated performance scores between Qwen2.5-7B and GPT-4 are very close on multiple models, demonstrating the general adaptability of our evaluation methods on multiple LLMs. This makes the evaluation process of HIS-Bench more easily accessible and reproducible.

\input{tables/supp-evaluator_reproduce}

\section{More HIS-Bench Data Samples}
\label{sec:supp-hisbench samples}

\begin{figure*}[htbp]
    \centering
    \includegraphics[width=0.95\linewidth]{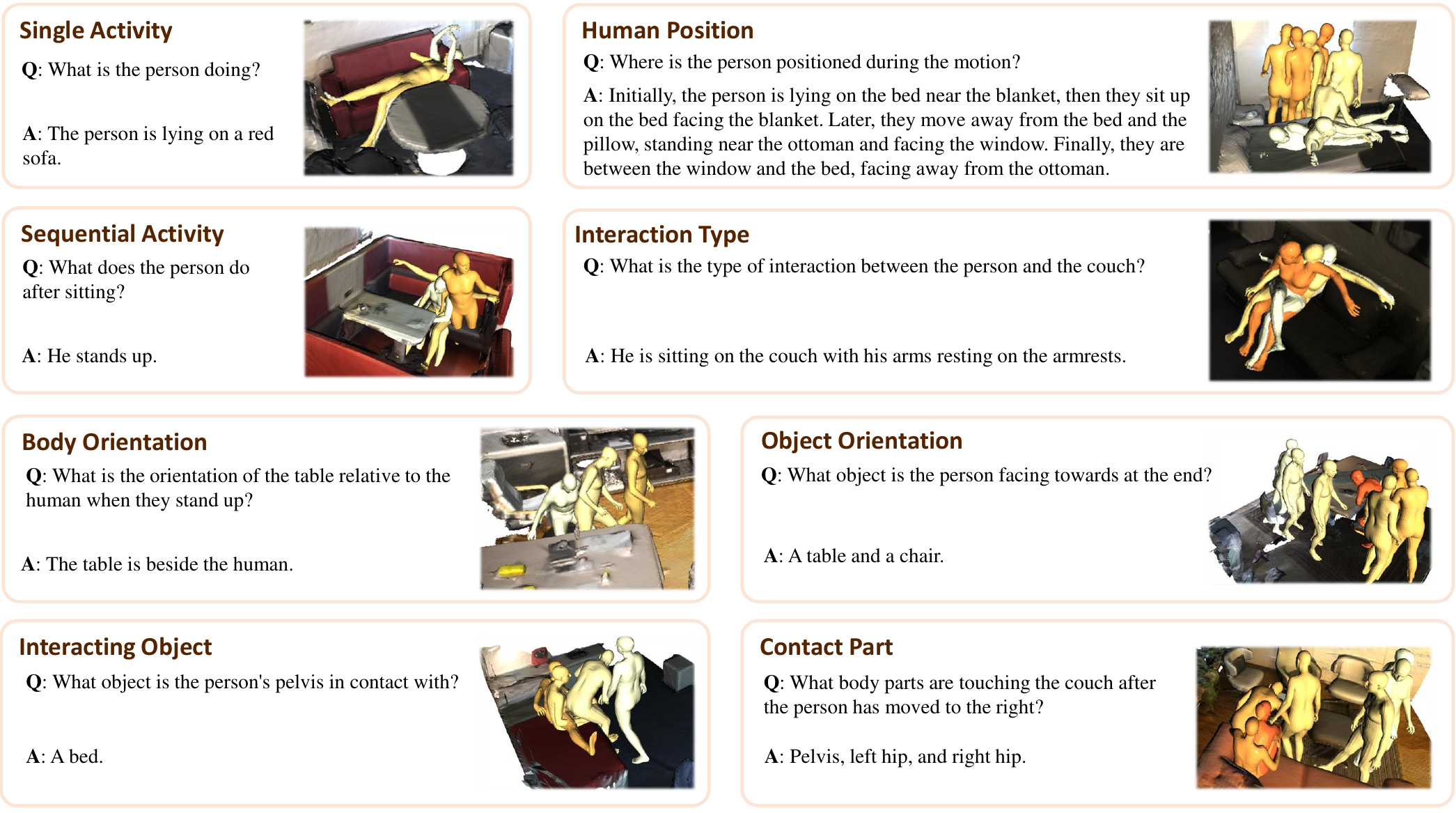}
    \caption{Examples of HIS-Bench, under the \textbf{`basic perception'} core ability, with 8 sub-tasks in total.}
    \label{fig:hisbench-basic-example}
\end{figure*}

\begin{figure*}[htbp]
    \centering
    \includegraphics[width=0.95\linewidth]{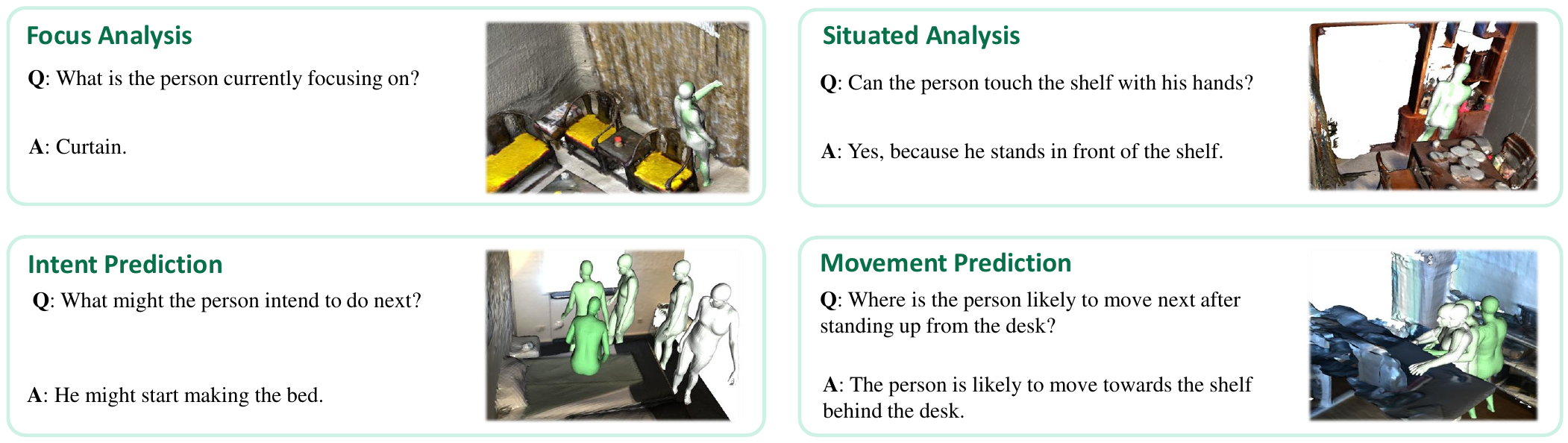}
    \caption{Examples of HIS-Bench, under the \textbf{`complex reasoning'} core ability, with 4 sub-tasks in total.}
    \label{fig:hisbench-reason-example}
\end{figure*}

\begin{figure*}[htbp]
    \centering
    \includegraphics[width=0.95\linewidth]{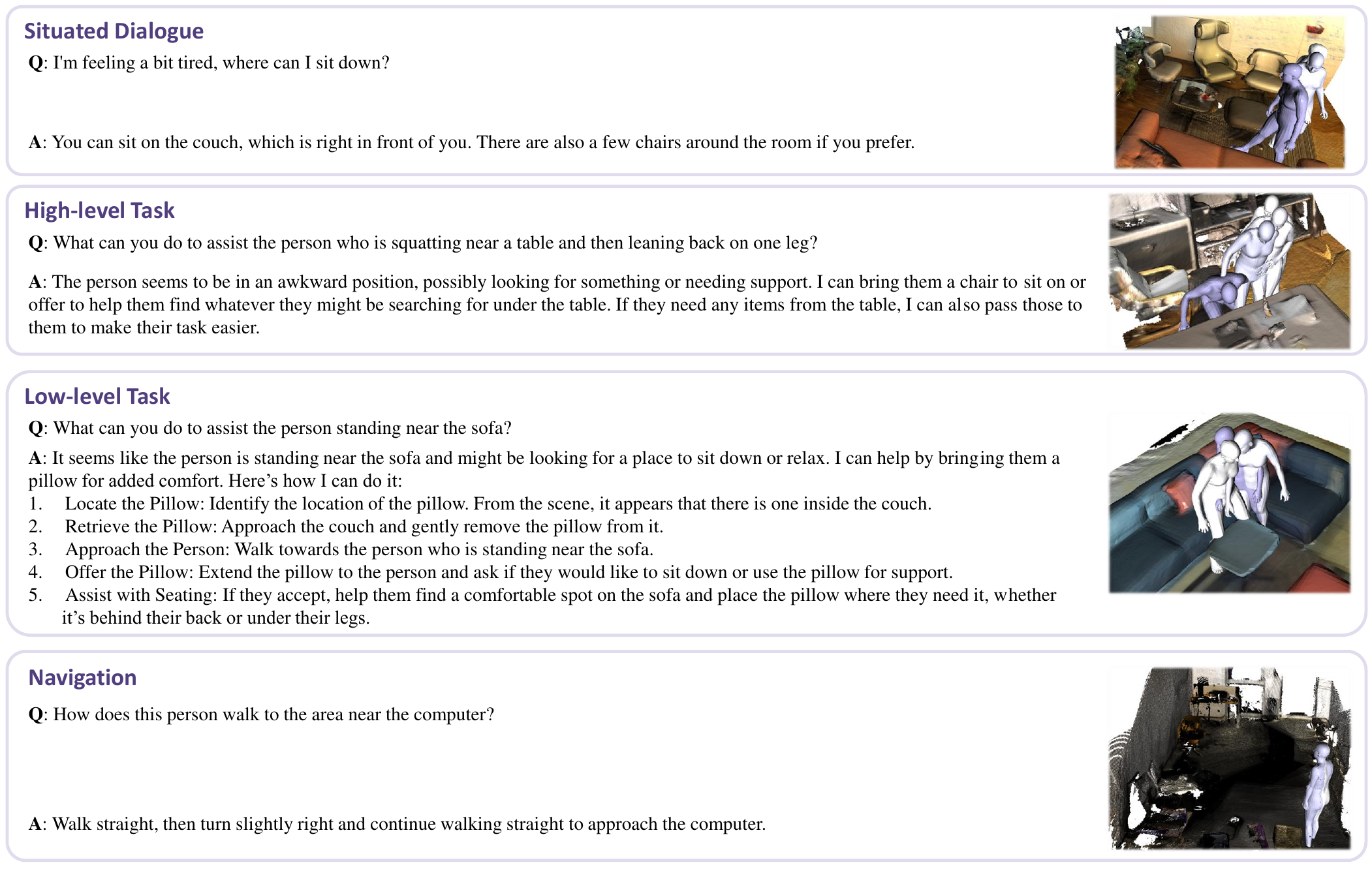}
    \caption{Examples of HIS-Bench, under the \textbf{`embodied applications'} core ability, with 4 sub-tasks in total.}
    \label{fig:hisbench-embodied-example}
\end{figure*}

In Fig.~\ref{fig:hisbench-basic-example}-\ref{fig:hisbench-embodied-example}, we present more data samples of HIS-Bench.~\cref{fig:hisbench-basic-example} showcases samples under the core ability of basic perception, covering a total of 8 sub-tasks.~\cref{fig:hisbench-reason-example} shows examples under the core ability of complex reasoning, covering 4 sub-tasks.~\cref{fig:hisbench-embodied-example} shows examples under the core ability of embodied applications, which contains 4 sub-tasks.

\section{Experimental Details}
\label{sec:supp-experiment details}

In this part, we elaborate on the experimental details, including the instruction templates (Appendix \ref{sec:instruction_templates}) and training data (Appendix \ref{sec:training_data}) for HIS-GPT, the implementation details for HIS-QA baseline models (Appendix \ref{sec:details_HIS_QA_baselines}), and the prompts designed for the GPT-assisted evaluation of HIS-Bench (Appendix \ref{sec:prompt_evaluation}).

\subsection{Instruction Templates}
\label{sec:instruction_templates}

Our training tasks include various input visual modalities: scene only, human motion only, and human-scene inputs. For each set of input modalities, we design a specific instruction template as follows:
 
\begin{itemize}
    \item Scene only: ``Examine the indoor scene. Object information in the scene: [\texttt{REPLACE}].'' 
    \item Motion only: ``Examine the human motion sequence. Motion sequence: [\texttt{REPLACE}].'' 
    \item Scene and Motion: Stage1:  ``Examine the indoor scene and a human motion sequence in the scene. Object information in scene: [\texttt{REPLACE}]. Motion sequence in scene:[\texttt{REPLACE}].''  Stage2: ``The conversation centers around an indoor scene and a human motion sequence. Object information in scene: [\texttt{REPLACE}]. Motion sequence: [\texttt{REPLACE}]. Based on the provided information, give an accurate answer to the following question raised by the user:'' .
\end{itemize}
Here \texttt{[REPLACE]} is replaced by scene/motion embeddings before feeding into LLM.

\input{tables/training-data}

\subsection{Training Data}
\label{sec:training_data}

In~\cref{tab:training data}, we provide a comprehensive list of the data used during the training of HIS-GPT. Specifically, for stage 1 of  Modality Alignment, we incorporate a total of 33.4k HIS caption data from HUMANISE~\cite{wang2022humanise} and TRUMANS~\cite{jiang2024scaling}, which are generated using our text annotation pipeline. Additionally, to facilitate the alignment of each modality in HIS data, we use 1.5k scene captions from SceneVerse~\cite{jia2024sceneverse} and 21k human motion captions from HumanML3D~\cite{guo2022generating}, respectively. For stage 2  of Instruction Tuning, we synthesize 700k diverse QA data using HUMANISE and TRUMANS datasets.

\begin{figure*}
    \centering
    \includegraphics[width=0.9\linewidth]{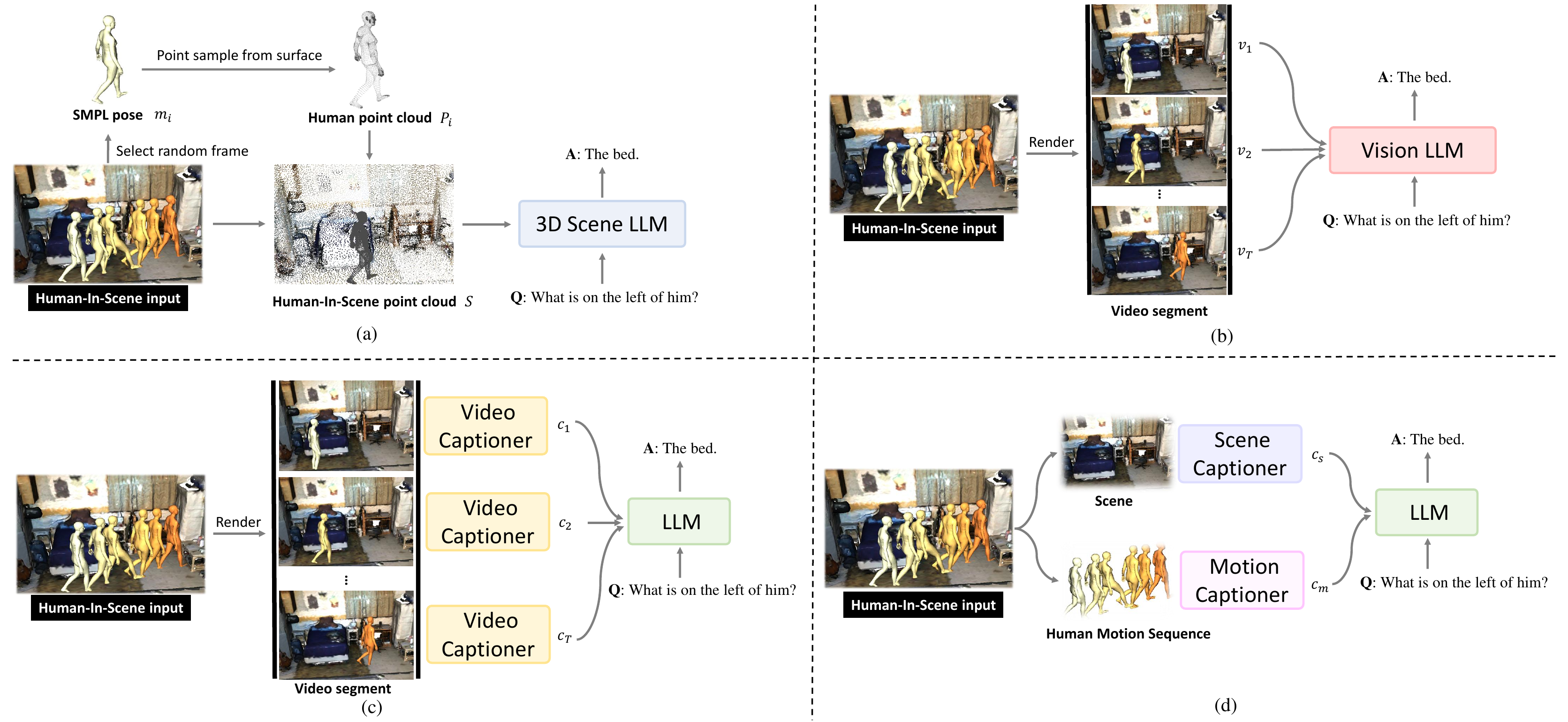}
    \caption{Illustrations of the HIS-QA baselines. (a) 3D Scene LLMs. (b) Vision LLMs. (c) LLMs w/ Frame Captions. (d) LLMs w/ Scene\&Motion Captions.}
    \label{fig:hisqa-baselines}
\end{figure*}

\subsection{Detailed Implementation of HIS-QA Baselines}
\label{sec:details_HIS_QA_baselines}

We have developed a set of baseline models for HIS-QA, as shown in \cref{fig:hisqa-baselines}. In this part, we provide the implementation details of these baselines. 

\noindent\textbf{3D Scene LLMs.} \ Current 3D scene LLMs are not capable of receiving sequential human motion input in SMPL format. Therefore, as shown in~\cref{fig:hisqa-baselines}(a), we select one frame $m_i$ from the motion sequence $\mathcal{M}$, densely sample a set of vertices $\mathcal{P}_i$ from the SMPL-fitted 3D human mesh, and feed these vertices into the 3D scene LLM together with the scene mesh to obtain an answer, \ie $\hat{\mathcal{A}}=\mathrm{f_{3D}}\left( \left[ \mathcal{S}, \mathcal{P}_i \right], \mathcal{Q} \right)$.

\noindent\textbf{Vision LLMs.} \ Existing vision LLMs excel at understanding image and video input, but cannot directly process 3D input. Therefore, as shown in~\cref{fig:hisqa-baselines}(b),we render the 3D HIS data into video segments $\mathcal{V}=\{v_i\}_{i=1}^T$, and leverage image or video LLMs to perform HIS-QA task, \ie $\hat{\mathcal{A}}=\mathrm{f_{VLM}}(\mathcal{V}, \mathcal{Q})$.

\noindent\textbf{LLMs w/ Frame Captions.} \ As shown in~\cref{fig:hisqa-baselines}(c), by leveraging powerful image captioners, we first generate frame-level captions $\mathcal{C}=\left\{ c_i \right\}_{i=1}^{T}$ from $T$ frames evenly sampled from $\mathcal{V}$. Then we use an LLM to answer the HIS-QA question, conditioned on the frame captions, \ie $\hat{\mathcal{A}}=\mathrm{LLM} \left(\mathcal{C}, \mathcal{Q} \right)$.

\noindent\textbf{LLMs w/ Scene\&Motion Captions.} \ As shown in~\cref{fig:hisqa-baselines}(d), we also try to derive linguistic information of HIS data by separately using scene and motion captioners to generate captions for 3D scene and 3D human motion sequence. Then we feed the scene caption $c_s$ and motion caption $c_m$ into an LLM, obtaining the answer $\hat{\mathcal{A}}=\mathrm{LLM} \left(c_s, c_m, \mathcal{Q} \right)$.

\begin{figure}
    \centering
    \includegraphics[width=1.0\linewidth]{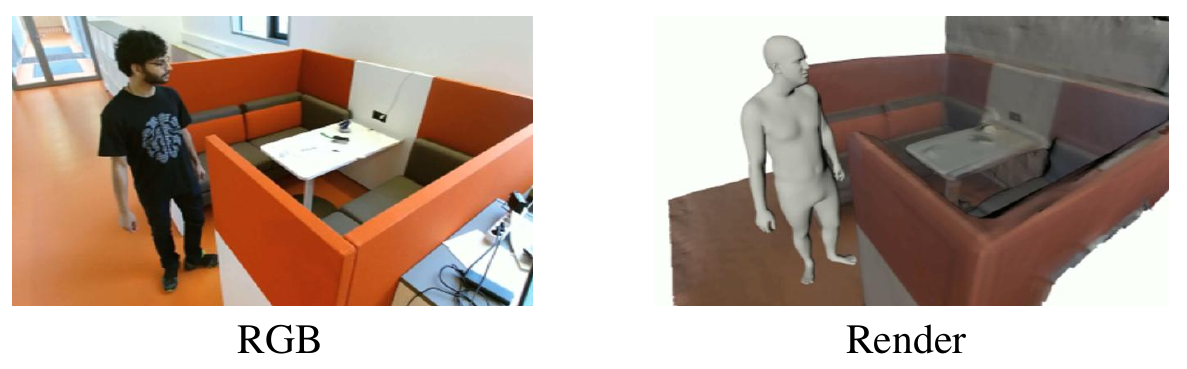}
    \caption{A comparison between the RGB third-person video frame (left) and the rendered video frame from raw 3D HIS data (right).}
    \label{fig:rgb_render_compare}
\end{figure}

\subsection{Prompt for GPT-assisted Evaluation}
\label{sec:prompt_evaluation}

\begin{figure*}
    \centering
    \includegraphics[width=0.95\linewidth]{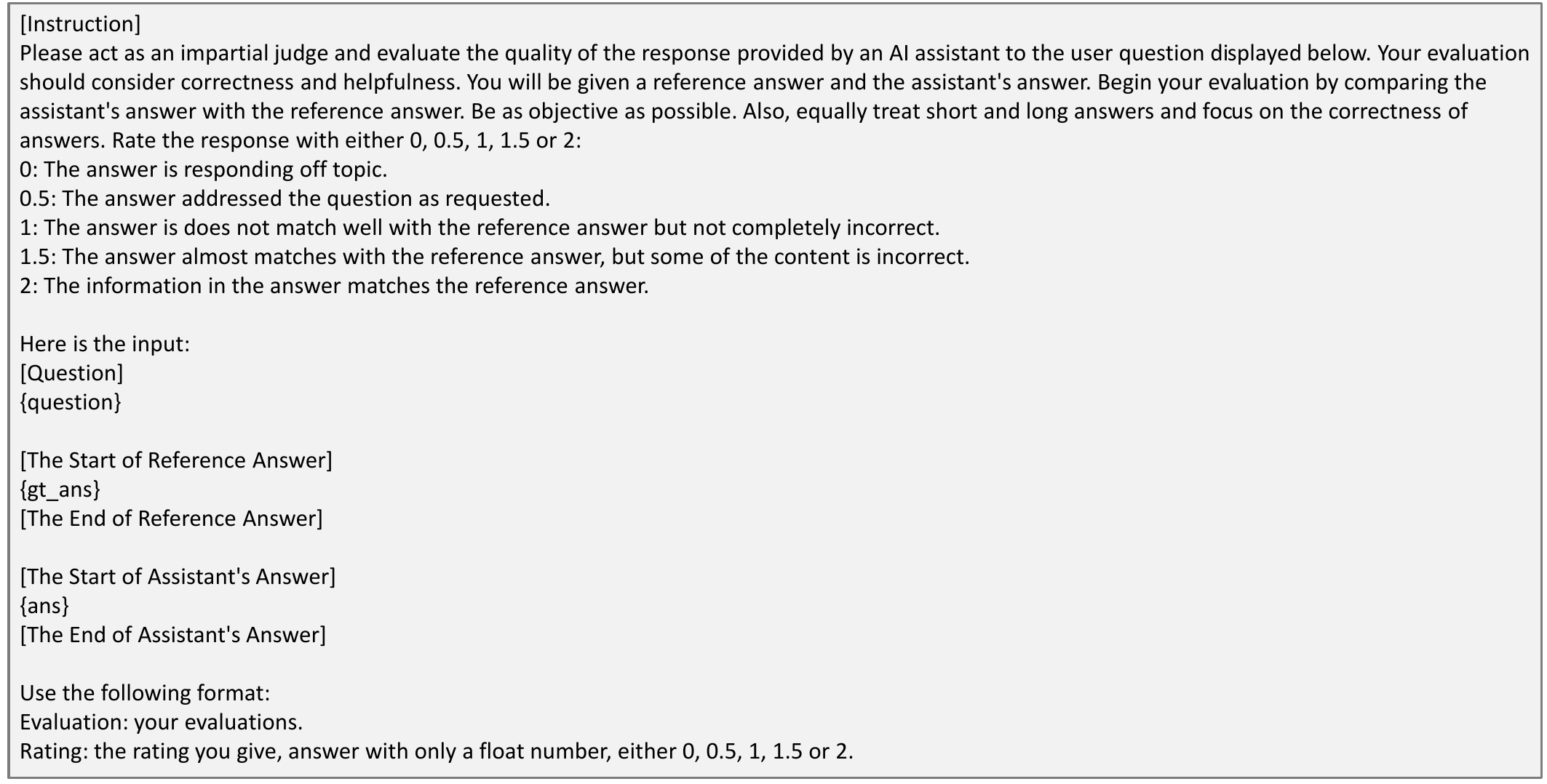}
    \caption{The prompt used for GPT-4 evaluation of HIS-Bench.}
    \label{fig:evaluate_prompt}
\end{figure*}

In~\cref{fig:evaluate_prompt}, we show the prompts used for the GPT-assisted evaluation of HIS-Bench. The prompt template follows~\cite{cheng2024egothink}.

\input{tables/supp-baseline_finetune}

\section{More Quantitative Results}
\label{sec:supp-quantitative}

\subsection{Results of Fine-tuned Baselines}

Since the majority of our compared baselines could not directly receive the input format of HIS-GPT training data, we test their results on HIS-Bench in a zero-shot manner. To make a fairer comparison, we convert the training corpus of HIS-GPT into compatible formats for the baseline methods, and fine-tune these baselines with the training data. We report the results of these fine-tuned models on HIS-Bench. As shown in~\cref{tab:baseline_finetune}, although being fine-tuned, the performance of baseline methods still largely lag behind HIS-GPT, showing the necessity of raising the HIS-GPT framework to process 3D human and scene modalities in a unified approach, which is aware to 3D natures and human-scene interactions.

\subsection{Ablations on Loss Weight}

\begin{figure*}
    \centering
    \includegraphics[width=0.8\linewidth]{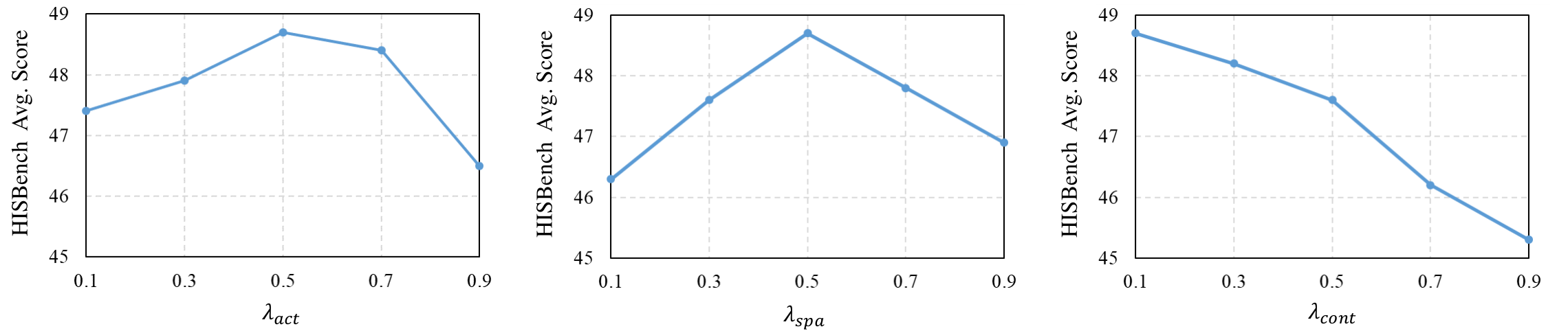}
    \caption{Ablations on the loss weight selection.}
    \label{fig:ablation-loss-weight}
\end{figure*}

In~\cref{fig:ablation-loss-weight}, we conduct ablations on the weights of different loss components in the training loss of HIS-GPT stage 1. From the results, we observe that the optimal weights for the three losses are $\lambda_{act}=0.5$, $\lambda_{spa}=0.5$ and $\lambda_{cont}=0.1$, respectively. %This validates our choices of hyper-parameters in HIS-GPT training. 
Moreover, when the loss weights are too large, the performance on HIS-Bench declines. We argue that an excessively large weight of auxiliary tasks can interfere with the supervision of autoregressive objectives during LLM training.

\subsection{Ablations on LLM Tuning Strategy}

For HIS-GPT training, to preserve the instruction-following and generalization ability of the LLM backbone, we keep the entire LLM frozen during all training stages. \cref{tab:ablation-llm-tuning} compares the performance of freezing the entire LLM with using LoRA~\cite{hu2022lora} for LLM tuning. The results show that keeping the LLM frozen achieves better performance than adopting LoRA. This suggests that maintaining LLM frozen is crucial for achieving satisfactory performance in HIS understanding, likely because the inherent reasoning and generalization capabilities of the pre-trained LLMs are well preserved. Notably, performance significantly declines when using LoRA for modality alignment (Stage 1), as it may cause the LLM to overfit to the caption data, thereby losing instruction following and complex reasoning abilities to some extent.
% Through case studies, we find that using LoRA could cause the model to overfit to the most dominant language contents in training data, thus hindering the evaluation performance. Further increasing the diversity of training corpus could be a solution to this problem.

\input{tables/supp-ablation-llm-tuning}

\input{tables/supp-ablation-rgb}

\subsection{Video Format for Vision LLM Evaluation}

For the evaluation of vision LLM baselines, we choose to render the HIS data into third-person videos and input them into the vision LLMs. This raises a concern that the domain gap between the rendered video and the common RGB input of vision LLMs could affect their performance. To explore this issue, we experiment with using the RGB video provided in the original data source of HIS-Bench for evaluating vision LLMs. A comparison between the RGB video and our rendered video is shown in~\cref{fig:rgb_render_compare}. As shown in ~\cref{tab:ablation-rgb}, using rendered video does not weaken the model's performance on HIS-Bench. In fact, on both Qwen-vl-max~\cite{bai2023qwen} and GPT-4o~\cite{hurst2024gpt}, rendered videos even exhibit higher performance than RGB videos. We hypothesize that the reason is that rendered videos from 3D data can more clearly present the spatial relationships between human and objects in the HIS data.

\section{More Qualitative Results}
\label{sec:supp-qualitative}

\begin{figure*}[htbp]
    \centering
    \includegraphics[width=0.95\linewidth]{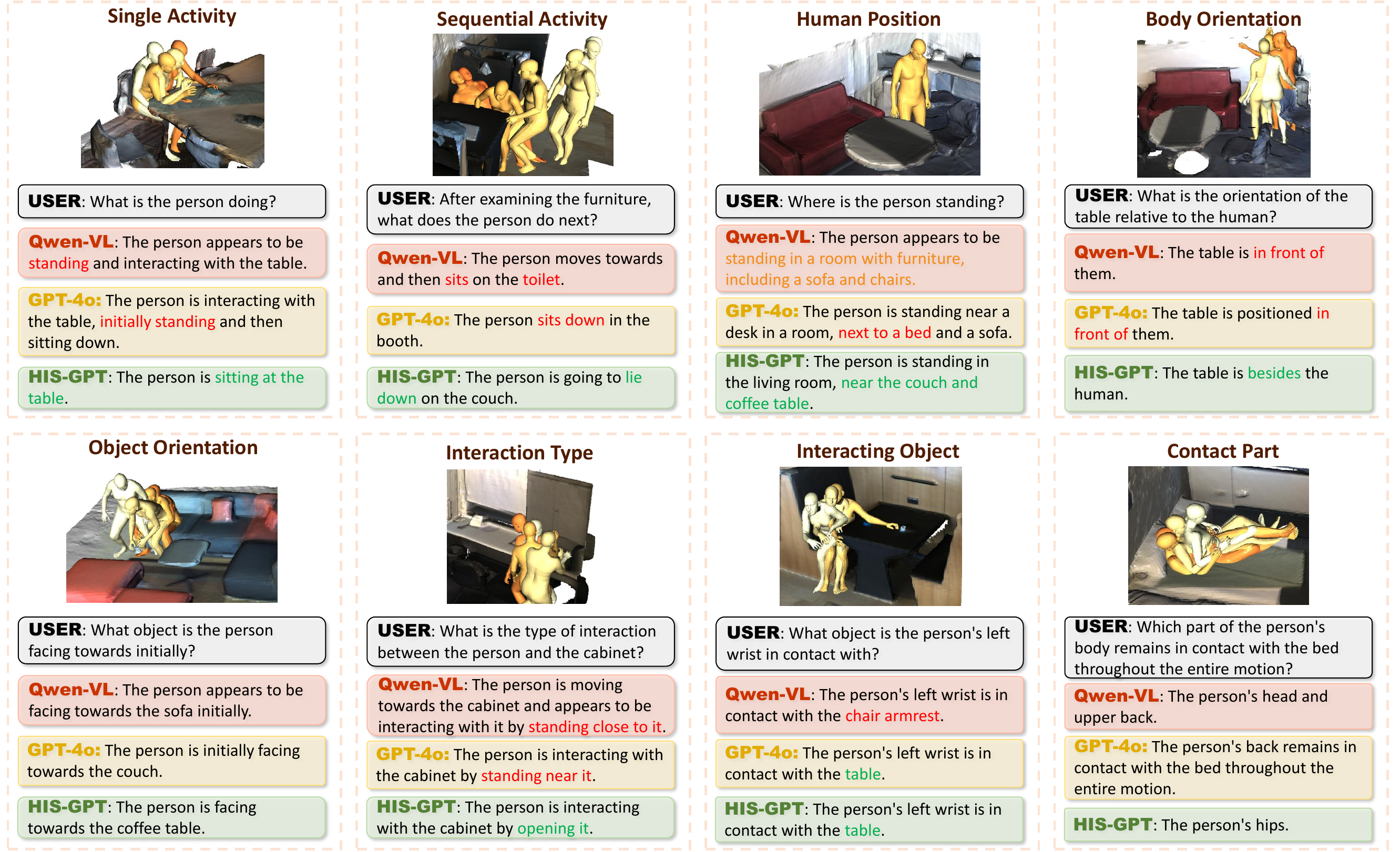}
    \caption{Qualitative examples of HIS-GPT on `basic perception' sub-tasks in HIS-Bench, compared with Qwen-vl-max and GPT-4o.}
    \label{fig:qualitative_basic}
\end{figure*}

\begin{figure*}[htbp]
    \centering
    \includegraphics[width=0.95\linewidth]{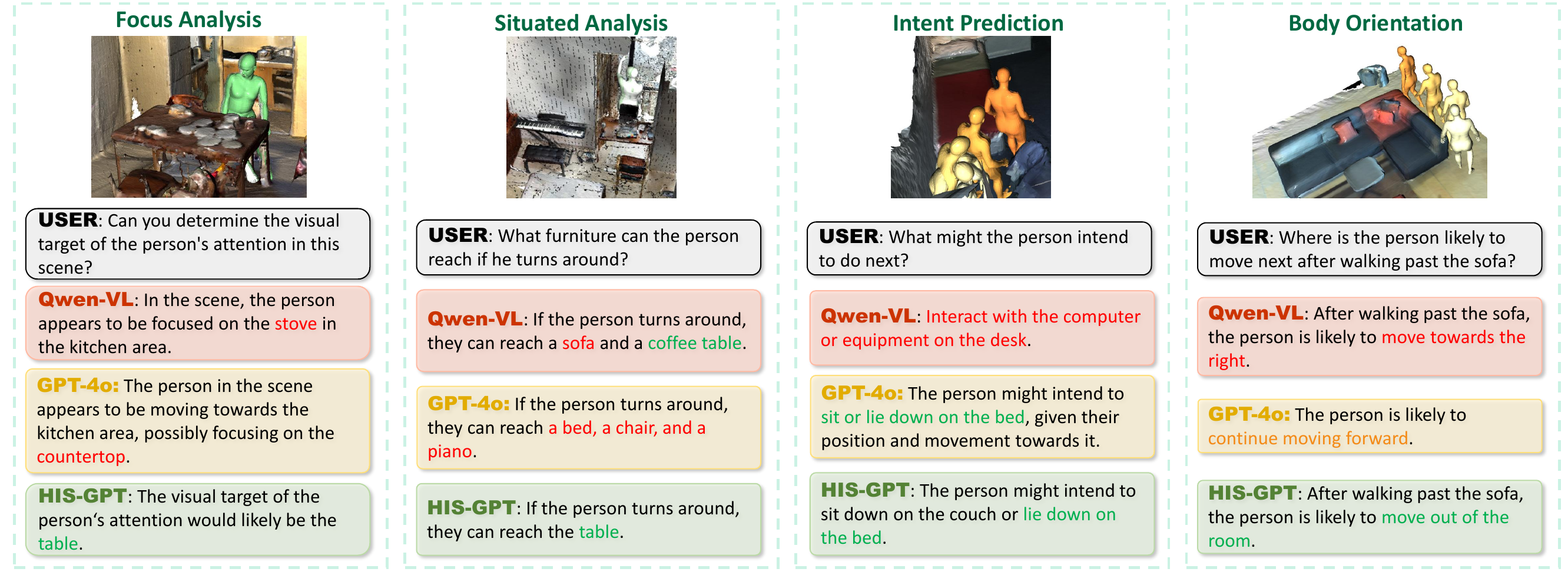}
    \caption{Qualitative examples of HIS-GPT on `complex reasoning' sub-tasks in HIS-Bench, compared with Qwen-vl-max and GPT-4o.}
    \label{fig:qualitative_complex}
\end{figure*}

\begin{figure*}[htbp]
    \centering
    \includegraphics[width=0.95\linewidth]{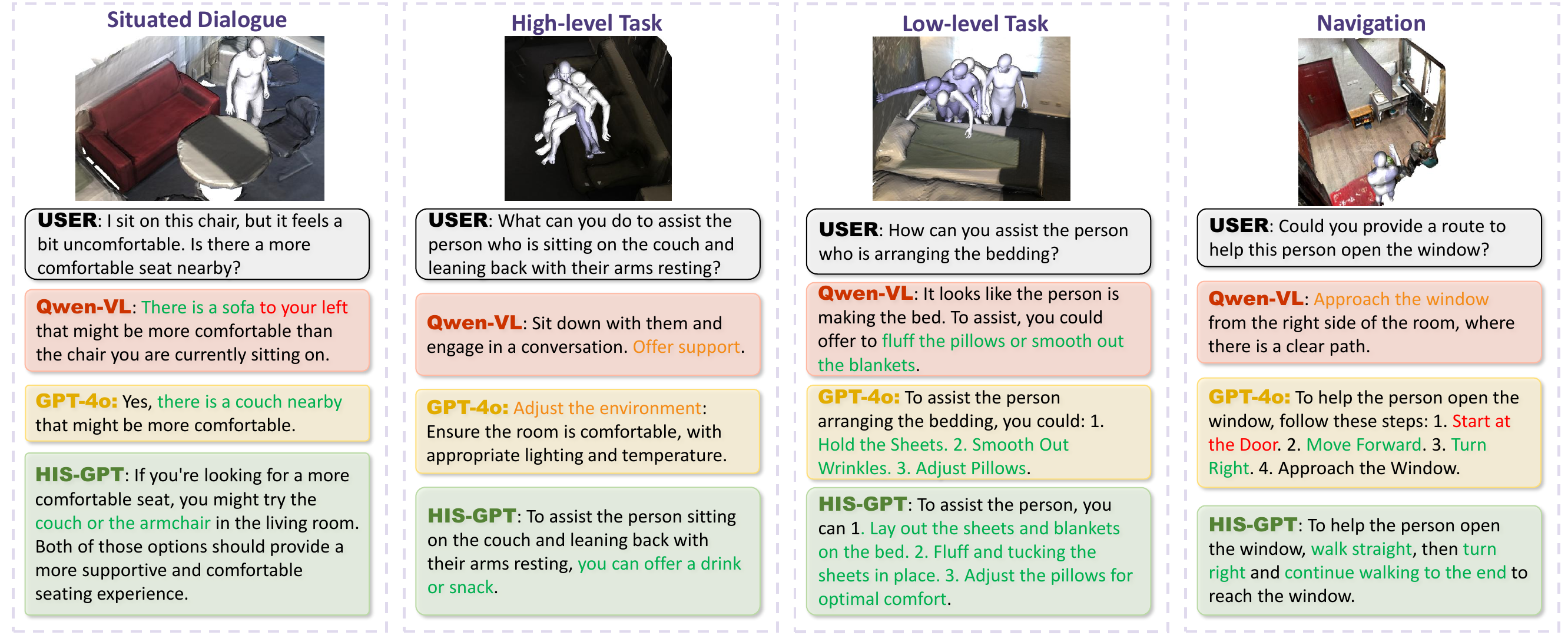}
    \caption{Qualitative examples of HIS-GPT on `embodied application' sub-tasks in HIS-Bench, compared with Qwen-vl-max and GPT-4o.}
    \label{fig:qualitative_embodied}
\end{figure*}

In Fig.~\ref{fig:qualitative_basic}-\ref{fig:qualitative_embodied}, we present more qualitative results of HIS-GPT on HIS-Bench. Compared with existing vision-language baselines, HIS-GPT demonstrates a significant advantage in basic perception, complex reasoning, and embodied application abilities.

\newpage

%\begin{figure*}[!ht]

%% file: tables/supp-userstudy.tex
\begin{table}[]
\centering
    \caption{User study on the quality of HIS-Bench. We show the average score of all samples in each dimension, graded by three annotators separately.}
\begin{adjustbox}{width=\linewidth,center}
\tablestyle{6pt}{1.0}
\begin{tabular}{c|cccc}
\toprule
    Benchmark & Answerability & Clarity & Correctness & Difficulty \\
\midrule
    HIS-Bench & 4.74 & 4.72 & 4.35 & 3.14 \\
    OpenEQA~\cite{majumdar2024openeqa} & 4.72 & 4.80 & 4.64 & 3.23 \\
\bottomrule
\end{tabular}
\end{adjustbox}
\label{tab:user_study}
\end{table}

%% file: tables/supp-evaluator_reproduce.tex
\begin{table}[]
\centering
    \caption{The consistencies of HIS-Bench average scores on multiple baseline models and HIS-GPT, by using GPT-4~\cite{luo2024m3gpt} and Qwen2.5-7B~\cite{bai2025qwen2} as evaluators, respectively.}
\begin{adjustbox}{width=\linewidth,center}
\tablestyle{6pt}{1.0}
\begin{tabular}{c|ccccc}
\toprule
    Evaluator & Chat-Scene & GPT-4o & LLaVAOV+GPT4 & LL3DA+AvatarGPT+GPT4 & \textbf{HIS-GPT} \\
\midrule
    GPT-4 & 8.2 & 31.3 & 17.9 & 5.0 & \textbf{48.7} \\
    Qwen2.5 & 7.4 & 29.6 & 16.9 & 4.4 & \textbf{46.1} \\
\bottomrule
\end{tabular}
\end{adjustbox}
\label{tab:evaluator_reproduce}
\end{table}

%% file: tables/training-data.tex
\begin{table}
\centering
\caption{Statistics of training data. We generate captions and QA pairs with both scene and motion input, serving as training corpus for HIS tasks. We also collect existing scene and motion caption data to facilitate modality alignment in stage 1.}
\begin{adjustbox}{width=\linewidth,center}
\tablestyle{8pt}{1.0}
\begin{tabular}{l|l|cccc}
    \toprule
     Stage & Datasets & scene & motion & type & \#pairs \\
     \midrule
     \multirow{4}{*}{Stage 1} & HUMANISE~\cite{wang2022humanise} & \Checkmark & \Checkmark & Caption & 24k \\
                              & TRUMANS~\cite{jiang2024scaling} & \Checkmark & \Checkmark & Caption & 9.4k \\
                              & SceneVerse~\cite{jia2024sceneverse} & \Checkmark & \XSolidBrush & Caption & 1.5k \\
                              & HumanML3D~\cite{guo2022generating} & \XSolidBrush & \Checkmark & Caption & 21k \\
     \midrule
     \multirow{2}{*}{Stage 2} & HUMANISE~\cite{wang2022humanise} & \Checkmark & \Checkmark & QA & 491k \\
                              & TRUMANS~\cite{jiang2024scaling} & \Checkmark & \Checkmark & QA & 209k \\
     \bottomrule
\end{tabular}
\end{adjustbox}
\label{tab:training data}
\end{table}

%% file: tables/supp-baseline_finetune.tex
\begin{table}[]
\centering
    \caption{The average scores of zero-shot and fine-tuned baseline methods on HIS-Bench. The fine-tuning data is consistent with HIS-GPT training data, as listed in~\cref{tab:training data}. HIS-GPT does not have zero-shot version since it is trained with HIS data from scratch.}
\begin{adjustbox}{width=\linewidth,center}
\tablestyle{6pt}{1.0}
\begin{tabular}{c|cccc}
\toprule
    Eval Model & Chat-Scene & LLaVAOV & LLaVAOV+GPT4 & \textbf{HIS-GPT} \\
\midrule
    Zero-shot & 8.2 & 14.2 & 17.9 & - \\
    Fine-tuned & 13.4 & 19.7 & 20.8 & \textbf{48.7} \\
\bottomrule
\end{tabular}
\end{adjustbox}
\label{tab:baseline_finetune}
\end{table}

%% file: tables/supp-ablation-llm-tuning.tex
\begin{table}[]
\centering
    \caption{Ablations on the LLM tuning strategy of HIS-GPT.}
\begin{adjustbox}{width=\linewidth,center}
\tablestyle{3pt}{1.0}
\begin{tabular}{cc|ccccccc|c}
\toprule
    \multicolumn{2}{c|}{\textbf{LoRA}} & \multicolumn{8}{c}{\textbf{HIS-Bench}} \\
    Stage 1 & Stage 2 & Act. & Spa. & HoI. & Ana. & Pre. & Dia. & Pla. & \textbf{Avg.} \\
\midrule
    \XSolidBrush & \Checkmark & 40.3 & \textbf{43.8} & 54.5 & 32.0 & \textbf{51.5} & \textbf{55.0} & 49.2 & 46.6 \\
    \Checkmark & \Checkmark & 40.5 & 39.7 & 42.7 & 27.8 & 49.5 & 46.5 & 26.9 & 38.1 \\
\midrule
    \XSolidBrush & \XSolidBrush & \textbf{44.6} & 42.1 & \textbf{55.5} & \textbf{41.0} & 50.3 & 53.2 & \textbf{53.9} & \textbf{48.7} \\
\bottomrule
    
\end{tabular}
\end{adjustbox}
\label{tab:ablation-llm-tuning}
\end{table}

%% file: tables/supp-ablation-rgb.tex
\begin{table}[]
\centering
    \caption{Ablations on the input video type for Vision LLM evaluation. `Render' denotes using the rendered videos from original 3D scene point cloud and 3D human SMPL data. `RGB' denotes using the recorded video data in these HIS datasets, which is filmed by RGB cameras in third-person view.}
\begin{adjustbox}{width=\linewidth,center}
\tablestyle{3pt}{1.0}
\begin{tabular}{c|c|ccccccc|c}
\toprule
    \multirow{2}{*}{\textbf{Model}} & \multirow{2}{*}{\textbf{Video type}} & \multicolumn{8}{c}{\textbf{HIS-Bench}} \\
    && Act. & Spa. & HoI. & Ana. & Pre. & Dia. & Pla. & \textbf{Avg.} \\
\midrule
    \multirow{2}{*}{Qwen-vl-max} & RGB & 25.6 & 15.7 & 36.0 & 12.0 & 15.7 & 29.5 & 17.8 & 21.5 \\
    & Render & 28.7 & 17.6 & \textbf{37.1} & 13.4 & 14.5 & 33.0 & 22.1 & 23.5 \\
\midrule
    \multirow{2}{*}{GPT-4o} & RGB & 29.5 & 17.1 & 35.8 & 33.5 & 13.9 & 33.0 & \textbf{35.9} & 28.3 \\
    & Render & \textbf{30.2} & \textbf{25.8} & 36.6 & \textbf{35.5} & \textbf{20.5} & \textbf{36.5} & 34.8 & \textbf{31.3} \\
\bottomrule
    
\end{tabular}
\end{adjustbox}
\label{tab:ablation-rgb}
\end{table}